%% file: main_cvpr.tex
\documentclass[10pt,twocolumn,letterpaper]{article}

\usepackage[pagenumbers]{cvpr} % To force page numbers, e.g. for an arXiv version

\definecolor{cvprblue}{rgb}{0.21,0.49,0.74}
\usepackage[pagebackref,breaklinks,colorlinks,allcolors=cvprblue]{hyperref}
\usepackage{amsmath,bm}
\usepackage{wrapfig,lipsum,booktabs}
\usepackage[normalem]{ulem}
\input{math_commands.tex}

\usepackage{multirow}

\usepackage{listings}
\newcommand{\hoang}[1]{\textcolor{blue}{#1}}

\usepackage{lipsum}

\newcommand\blfootnote[1]{%
  \begingroup
  \renewcommand\thefootnote{}\footnote{#1}%
  \addtocounter{footnote}{-1}%
  \endgroup
}

\def\paperID{16913} % *** Enter the Paper ID here
\def\confName{CVPR}
\def\confYear{2025}

\title{Leveraging Hierarchical Taxonomies in Prompt-based Continual Learning}

\author{Quyen Tran\\
VinAI Research\\
{\tt\small v.quyentt15@vinai.io}
\and
Hoang Phan*\\
New York University\\
{\tt\small hvp2011@nyu.edu}
\and 
Minh Le*\\
VinAI Research\\
{\tt\small v.minhld12@vinai.io}
\and 
Tuan Truong\\
VinAI Research\\
{\tt\small v.tuantm27@vinai.io}
\and 
Dinh Phung\\
Monash University\\
{\tt\small Dinh.Phung@monash.edu}
\and 
Linh Ngo\\
Hanoi University of Science and Technology\\
{\tt\small linhnv@soict.hust.edu}
\and 
Thien Nguyen\\
University of Oregon\\
{\tt\small thienn@uoregon.edu}
\and 
Nhat Ho\\
The University of Texas at Austin\\
{\tt\small minhnhat@utexas.edu}
\and 
Trung Le\\
Monash University\\
{\tt\small trunglm@monash.edu}
}

\begin{document}

\maketitle

\blfootnote{*Hoang Phan and Minh Le contributed equally to this work}

% \begin{center}

% {\bf{\LARGE{Leveraging Hierarchical Taxonomies in Prompt-based Continual Learning}}}
  
% \vspace*{.2in}
% {\large{
% \begin{table*}[!ht]
%     \centering
%     \begin{tabular}{cccc}
% Quyen Tran$^{\diamond}$ & Minh Le$^{\diamond}$ & Tuan Truong$^{\diamond}$ &  Dinh Phung$^{\ddagger}$ \\
% Linh Ngo$^{\star}$ & Thien Nguyen$^{\star \star}$ & Nhat Ho$^{\dagger}$ &  
% Trung Le$^{\ddagger}$
% \end{tabular}

%     \vspace*{.2in}

% \begin{tabular}{ccc}
% VinAI Research$^{\diamond}$\\
% Monash University$^{\ddagger}$ \\
% Hanoi University of Science and Technolgy$^*$ \\
% Univesity of Oregon$^{**}$ \\
% The University of Texas at Austin$^{\dagger}$\\
% \end{tabular}

% \end{table*}

% }}

\begin{abstract}
\input{sec/0-abstract}

\end{abstract}
% \end{center}

\section{Introduction}
\input{sec/1-intro}

\section{Related Work}
\label{relatedW}
\input{sec/2-1-relatedwork}

\section{Preliminary Analysis}
\label{background}
\input{sec/2-2-background}

\section{Proposed Method}
\label{method}
\input{sec/3-framework}

\section{Experiments}
\label{exp}
\input{sec/4-exp}

\section{Conclusion and Discussion}
\label{conclusion}
\input{sec/5-conclusion}

\newpage
{
    \small
    \bibliographystyle{ieeenat_fullname}
    \bibliography{main}
}

% WARNING: do not forget to delete the supplementary pages from your submission 
% \input{sec/X_suppl}

\newpage
\appendix
\input{sec/6-appendix}

\end{document}

%% file: math_commands.tex
\usepackage{amsmath,amsfonts,bm}

\def\eqref#1{equation~\ref{#1}}
\def\1{\bm{1}}

\def\vx{{\bm{x}}}

\def\vz{{\bm{z}}}

\DeclareMathAlphabet{\mathsfit}{\encodingdefault}{\sfdefault}{m}{sl}
\SetMathAlphabet{\mathsfit}{bold}{\encodingdefault}{\sfdefault}{bx}{n}

%% file: sec/0-abstract.tex
Humans perceive the world as a series of sequential events, which can be hierarchically organized with different levels of abstraction based on conceptual knowledge. Drawing inspiration from this human cognition aspect, we explore how human tendencies to organize and connect information can enhance the training of deep learning models.  Specifically, we propose a novel approach to mitigate catastrophic forgetting by exploiting the relationships between continuously emerging class data. 
By building a hierarchical tree structure based on the expanding set of labels, we gain fresh insights into the data, identifying groups of similar classes could easily cause confusion.
Additionally, we delve deeper into the hidden connections between classes by exploring the original pretrained model behavior through an optimal transport-based approach. From these insights, we propose a novel regularization loss function that encourages models to focus more on challenging knowledge areas,  thereby enhancing overall performance. Experimentally, our method demonstrated significant superiority over state-of-the-art methods on various continual learning benchmarks. 
Our code is available at \url{https://anonymous.4open.science/r/HierC-089B/}.

% % Firstly, we uncover that the forgetting phenomenon in PCL models mainly occurs due to the introduction of new classes that may overlap with the old ones in the latent space. To address this, the model must effectively distinguish between old and new data, especially those with many similar characteristics. 

%% file: sec/1-intro.tex
% {\color{blue} Nhat: References?}
Continual Learning (CL) \citep{wang2024comprehensivesurveycontinuallearning, DBLP:conf/nips/Lopez-PazR17} is a research direction that focuses on realizing the human dream of creating truly intelligent systems, where machines can learn on the fly, accumulate knowledge, and operate in constantly changing environments as a human's companion. Despite the impressive capabilities of A.I systems, continual learning remains a challenging scenario due to the tendency to forget obtained knowledge when facing new ones, known as \textit{catastrophic forgetting} \citep{french1999catastrophic}. In dealing with this challenge, traditional CL methods often rely on storing past data for replaying during new tasks \cite{lopez2017gradient, buzzega2020dark}, which can raise concerns about memory usage and privacy. {Besides, prior work
shows that replay methods result in overfitting and poor generalization \cite{lopez2017gradient, verwimp2021rehearsal, rio2023studying}}.  To overcome these limitations, recent methods leverage the strong generalization ability of pretrained models \citep{han2021pre, jia2022visual} to solve sequences of CL tasks. A notable line of work is the prompt-based approach \citep{wang2022dualprompt, smith2022coda, li2024steering}, where a small set of learnable prompts is injected to pretrained backbones for adapting emerging tasks over time.

% achieving state-of-the-art performance on various continual learning benchmarks.

% P1: Human learning ---> CL ----> Traditional CL's drawback... privacy, require taskID when inference, forgetting.. .catastrophically.. --> Pretrained Continual Learning... 

While these prompt-based methods have demonstrably achieved impressive results, they only consider forgetting caused by changes of common parameters across tasks during learning \citep{wang2022learning_l2p, wang2022dualprompt} or the inherent mismatch between the chosen prompts at training and testing \citep{wang2023hierarchical, tran2023koppa, cprompt, smith2022coda}. In this work, we further complement these views by showing that the forgetting of old tasks can potentially come from the uncontrolled overlap between old and new emerging class representations in the latent space. That is, models become more confused in distinguishing these classes, resulting in performance degradation w.r.t previous tasks over time (i.e., forgetting). Furthermore, we find that existing methods only \textit{utilize limited information} from the training dataset and often \textit{treat classes equally} during training. Consequently, they overlook the opportunities to enhance the distinguishability of models, especially between the class representations of old and new tasks, and thus, hinder the models' ability to mitigate forgetting \hoang{\cite{french1991using, kaushikunderstanding}}.
% Therefore, in order to mitigate forgetting, we need to consider the the separability old and new classes of the model and mitigat 

In addition, we discover that human natural learning behavior has many valuable aspects, especially the habit of analyzing data, organizing them in a meaningful way, and finding connections between old and new knowledge \citep{schon1983reflective, bransford2000how, sweller1988cognitive, mayer2005cambridge}.
% , thereby improving the ability to understand, remember, and reproduce information. 
Inspired by these practices, we investigate the characteristics of common benchmark datasets, as well as the behavior of pretrained models. Our findings reveal that the data can always be categorized into consistent groups,  regardless of their arrival times.
Each group usually includes classes with similar semantic information that models may confuse and, thus, should be paid special attention to during training. Identifying these kinds of groups will benefit in improving model representation learning, as well as aiding in addressing catastrophic forgetting. 

Therefore, we propose a novel training strategy that constantly arranges emerging data in groups, following a \textit{tree-like taxonomy}. In particular, during training a new task, models are trained to distinguish all classes so far, especially focusing on classes \textit{within the same leaf group}. 
We observe that images belonging to concepts/labels within each of these \textit{leaf groups} share strong visual and semantic correlations, leading to overlap in the latent space, which compromises performance. By encouraging models to contrast these classes more distinctly, we can reduce the overlap of easily confused ones.
This strategy not only mitigates forgetting when new classes emerge but also consolidates domain-specific knowledge in each leaf group. 
% Additionally, to further exploit the hierarchical taxonomy, we propose a testing strategy that leverages the group's characteristics to better identify data samples, especially those near boundaries between different groups in the taxonomy. 
% Therefore, we propose a training strategy that constantly considers emerging data in groups, following a hierarchical tree-like taxonomy. In particular, during learning new tasks, a model always
% references information from old classes in the tree. Especially, the data in the same group will
% be applied a specific constraint to encourage the model to deeply understand the detailed knowledge contained in that group. This strategy not only helps the model avoid forgetting when new
% classes appear, but also consolidates the model’s knowledge in specific domains corresponding to
% each group. 
% In addition, based on human learning habits, we also find that individuals with better foundational knowledge will absorb knowledge faster and more effectively. Thus, we propose a novel optimal transport-based technique to further employ prior knowledge from pretrained models, where the behavior of this starting model will reveal another view of the relationships between classes. 
In addition, based on human learning habits, we find that individuals with stronger foundational knowledge absorb new information faster and more effectively. Thus, we propose an optimal transport-based technique to further utilize priori from pretrained models, where their initial behaviors will provide another perspective on the relationships between classes.

% Moreover, we propose a testing strategy where the characteristics of each group aid in identifying data samples, especially those located at
% the boundaries between different leaf groups of the taxonomy.

\textbf{Contribution.} We introduce a method named \emph{Leveraging Hierarchical \textbf{T}axonomies in Prompt-based \textbf{C}ontinual \textbf{L}earning} (TCL). Our main contributions are summarized as follows:

\begin{itemize}
    \item We conduct empirical analysis to suggest a new cause of catastrophic forgetting in prompt-based continual learning models, which inherently comes from the uncontrolled growth of incoming classes on the latent space.
    
    \item Inspired from Cognitive Science, we propose a novel approach to reduce forgetting by examining the relationships between data. By dynamically constructing label-based hierarchical taxonomies and leveraging the prior knowledge of pretrained models through an optimal transport-based approach, we can identify the challenging knowledge areas that require further focus during the sequence of tasks.

    % where classes with similar characteristics are consistently grouped.
    % In addition, to further extract the hidden connection between different classes of data, we offer another strategy ... 
    % Additionally, our testing strategy leveraging cluster information helped improve the model's prediction probability.
    
    % \item We empirically evaluate the effectiveness of our method against current state-of-the-art pretrained-based baselines across various continual learning benchmarks.
    \item Experimentally, our method demonstrated significant superiority over state-of-the-art methods on various continual learning benchmarks.
    
\end{itemize}

% \textbf{Organization.} The rest of the paper is structured as follows. In Section \ref{relatedW}, we present related work. Then in Section \ref{background}, we formulate the problem and introduce a new perspective to explain the cause of forgetting in prompt-based CL models. Section \ref{method} transitions from the motivation provided by Cognitive Science insights to the proposed training strategy, emphasizing the importance of leveraging relationships between class data. Section \ref{exp} presents the experimental results, and finally, we discuss the limitations and suggest future directions in Section \ref{conclusion}.

% In summary, our contributions... 
% ------------------------------------

% - Inspired by human behavior when learning new things, dive deep + connect with old stuffs 

% ---> Training: detail, find connection between information that has been learned and is being learned via manipulating latent space....

% + Take advantage of the prior (base) knowledge from pretrained model...

% ----> testing:Exploiting the relationships of learned information during inference + prior from base model...

% \subsection{Motivations}

% * CIFAR:
% HIDE:  92.61
% + tt (92.81)

% Ours: 93.98
% train only: 93.52 

% -------------------------\\
% * ImgNetR:
% HiDe: 75.06 
% + tt (75.22)

% Ours: 76.41
% train_only: 76.02

% -------------------------\\
% * CUB: 
% HiDE: 86.62
%  + tt (86.75)

% Ours: 88.02
% train_only: 87.80

%% file: sec/2-1-relatedwork.tex
% \textbf{Continual Learning.} Continual learning \citep{ DBLP:conf/nips/Lopez-PazR17, masana2022class, le2024mixtureexpertsmeetspromptbased} tackles the challenge of incrementally training a model on a sequence of tasks, while mitigating the risk of 
% The main problem of Continual Learning is finding the way to mitigate \textit{catastrophic forgetting} \citep{mccloskey1989catastrophic, nguyen2019toward}, where previously learned knowledge is lost as new tasks are introduced. 
\paragraph{Class Incremental Learning (CIL).}
This is one of the most challenging and widely studied CL scenarios \citep{van2019three, wang2023hierarchical}, where task identity is unknown during testing, and data of previous tasks is inaccessible during current training \citep{DBLP:journals/pami/MasanaLTMBW23, DBLP:conf/cvpr/RebuffiKSL17, DBLP:conf/cvpr/HouPLWL19, pmlr-v162-guo22g}.
% Most successful CIL methods rely on storing and rehearsing old training data, which introduces concerns around efficiency and privacy \citep{rebuffi2017icarl, wang2021memory, cha2021co2l}. Recently, approaches leveraging the strong generalization abilities of pretrained models have emerged, achieving state-of-the-art performance on various benchmark datasets \citep{zhou2023revisiting, wang2023isolation, mcdonnell2023ranpac}. 
This work follows the setting of CIL and proposes a novel approach for prompt-based CL models.
% , which take advantage of pretrained models as frozen backbones.  

\vspace{-3mm}
\paragraph{Prompt-based Continual Learning.}
This line of work exploits the power of pretrained backbone to quickly adapt to the sequence of downstream tasks by updating just a small number of parameters/ prompts for different tasks.
In initial work like \cite{wang2022learning_l2p, wang2022dualprompt, smith2022coda}, the absence of explicit training constraints often leads to feature overlapping between classes from different tasks. Therefore, recent methods employ some types of contrastive loss \citep{wang2023hierarchical, li2023steering} or utilize Vision Language models \citep{DBLP:conf/nips/WangHH22, DBLP:conf/wacv/NicolasCZADD24} to better separate features from tasks.
Nevertheless, they treat all classes equally during training, missing the opportunity to efficiently distinguish classes, which have many similarities and are easily confused. In this work, we propose a novel approach to reveal the relationships within data, allowing the model to recognize groups of these classes, and develop a deeper understanding of the respective knowledge areas, thereby effectively reducing forgetting.
\vspace{-3.8mm}
\paragraph{Using hierarchical label structures}
In Deep Learning, existing work \citep{DBLP:journals/pr/DimitrovskiKLD11, DBLP:conf/emnlp/ChalkidisFKMAA20, DBLP:conf/eccv/YiSGE22, DBLP:conf/naacl/LiuZYZ21} have considered using hierarchical label systems for efficient representation learning. However, these approaches require all labels in advance, which is not suitable for Continual Learning, and the trees are often fixed in both depth and width. Recent researches in CL \citep{DBLP:conf/iccv/LeeJCC23, DBLP:conf/coling/CaoZLXLK24} have considered hierarchical structures to either expand the label set to create new learning scenarios or manage memory buffers for rehearsal settings. However, these structures are still limited in representing information and especially cannot indicate the semantic relationships between labels for effective training.
% Differently, we propose to dynamically build a label-based taxonomy over time, where it is not necessary to provide all labels in advance for rehearsal-free models. Additionally, the tree construction will have the support of experts (i.e., human experts, or LLms), allowing for the acquisition of detailed information and the correlation between class data, thereby enhancing the learning of new knowledge and effectively reducing the forgetting of the old ones.
Alternatively, we propose to dynamically build a label-based taxonomy over time for rehearsal-free models, where it is not necessary to provide all labels in advance. Additionally, the tree construction will have the support of experts (i.e., human experts or LLMs), allowing for the acquisition of detailed information and the correlation between class data, thereby enhancing the learning of new knowledge and effectively reducing the forgetting of old information.

%% file: sec/2-2-background.tex
 In this part, we first present the problem formulation of prompt-based CL in Section~\ref{sec:prob_formulation}. Following this, we analyze the sources of forgetting in these methods in Section~\ref{sec:forgetting}.

\subsection{Problem formulation} \label{sec:prob_formulation}
% We consider the Continual learning setting in which a model is required to learn from a sequence of classification tasks. Each task $t \in \{1, ..., T\}$ has a training dataset $D_t$ with $n_t$ i.i.d. samples $\{(\boldsymbol{x}_t^i, y_t^i)\}_{i=1}^{n_t}$, and $T$ may be unknown or infinite. The model needs to learn these tasks sequentially, without replaying data from old tasks during training and without access to task ID when inference. 
% This approach is often known as \textit{rehearsal-free Continual Learning} \citep{smith2022coda, wang2023hierarchical}.

We consider the Class Incremental Learning setting \citep{Zhou_2024, lopez2017gradient, wang2023hierarchical}, where a model has to learn from a sequence of $T$ visual classification tasks without revisiting old task data during training or accessing task IDs during inference. Each task $t \in \{ 1, ..., T\}$ has a respective dataset $\mathcal{D}_t$, containing $n_t$ i.i.d. samples ${(\boldsymbol{x}_t^i, y_t^i)}_{i=1}^{n_t}$. 
% Let $D_c = (X_c,  Y_c)$ denote the data corresponding to the class label $c$.

In this work, we design our model as a composition of two components: \textit{a pretrained Vision Transformer (ViT) \citep{dosovitskiy2021vit} backbone} $f_\Phi$ and \textit{a classification head} $h_\psi$. That is, we have the model parameters $\theta = (\Phi, \psi)$. Similar to other existing prompt-based methods \cite{wang2023hierarchical, smith2022coda}, we incorporate into the pretrained ViT backbone a set of prompts $\bm{P}$. We denote the overall network after incorporating 
the prompts as $f_{\Phi, \bm{P}}$. 

\begin{figure*}[htbp]
    \centering
    
    \begin{subfigure}{0.23\textwidth}
        \centering
        % \vspace{-10mm}
        \resizebox{\textwidth}{!}{
        \begin{tabular}{lc}
            \toprule
              Dataset & Feature shift \\
            \midrule
            Split-Imagenet-R & 0\\
            Split-CIFAR-100 & 0 \\
            Split-CUB-200 & 0\\
            \bottomrule
        \end{tabular}}
        \vspace{2mm}
        \caption{Feature shift corresponding to $\mathcal{D}_1$ after learning sequence of tasks.}
        \label{tab:shift}
        \vspace{13mm}
    \end{subfigure}
    \hfill
    \begin{subfigure}{0.35\textwidth}
        \centering
        \includegraphics[width=\linewidth, height=4.9cm]{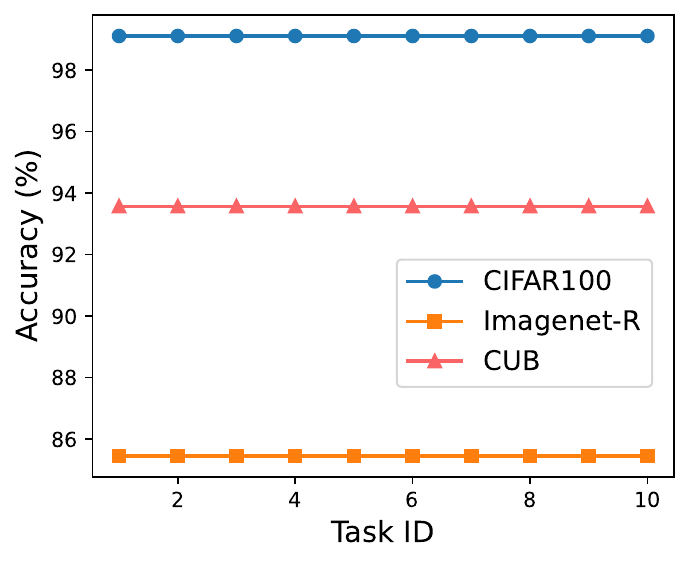}
        \caption{\textit{"Within task accuracy"} on $\mathcal{D}_1$.}
        \label{fig:wt_a}
    \end{subfigure}
    \hfill
    \begin{subfigure}{0.35\textwidth}
        \centering
        \includegraphics[width=\linewidth]{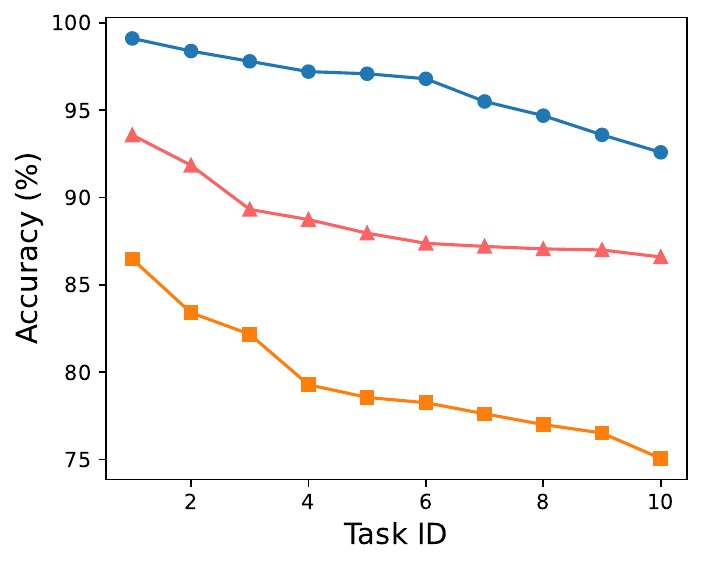}
        \caption{\textit{Model accuracy} on $\mathcal{D}_1$}
        \label{fig:true_a}
    \end{subfigure}
    % \vspace{-3mm}
    \caption{\textbf{Empirical study about forgetting.} We setup the experiment to eliminate factors \textbf{\textit{(I)}} and \textbf{\textit{(II),}} which result in feature shift after learning the sequence of tasks - \textbf{Table (a)}. Therefore, the \textit{"within task accuracy"} on $\mathcal{D}_1$, using classification head $s_1(\bm{x})$ to classify classes within task 1 only, remains over time - \textbf{Figure (b)}. However, when using head $h_\psi$ to classify all classes observed so far, the \textit{model accuracy} on $\mathcal{D}_1$ decrease significantly - \textbf{Figure (c)}, suggesting that besides \textit{feature shift, there are other factors that lead to forgetting}.}
    \label{fig:insight}
    \vspace{-2mm}
\end{figure*}

\subsection{Forgetting in Prompt-based Methods} \label{sec:forgetting}
In Continual Learning, \textit{forgetting} refers to the phenomenon in which performance on previously learned tasks decreases over time. Current prompt-based CL methods, which leverage the power of pretrained models, attribute forgetting either to \textbf{(I)} changes in backbone parameters when using the common prompt pool $\bm{P}$ for all tasks \citep{wang2022learning_l2p, wang2022dualprompt} or to \textbf{(II)} the inherent mismatch between the models used at training and testing,
% Specifically, let $\hat{t}({\bm{x}})$ and $t({\bm{x}})$ as the chosen promptID and the ground-truth promptID for $\bm{x}$, respectively. We may have the promptID $\hat{t} \neq t$, where $\hat{t}$ is predicted by pretrained backbone $f_\Phi$, thus there are chances that
% $f_{\hat{\theta}_t} = f_{\Phi, \bm{P}_{\hat{t}}} \neq f_{{\theta}_t} = f_{\Phi, \bm{P}_{{t}}}
% $, 
as discussed and analyzed in \cite{cprompt, tran2023koppa}.
In this work, we complement these views by providing empirical studies to offer new insights about the reason for forgetting in this type of model, which arises from overlapping between old and new emerging class representations. 
% {\color{blue}Tuan: we can summarize our conclusions here, what are those insights? "Our investigations suggest/show/conclude that...", as far as I understand, it's about the overlapping phenomena of new classes. Also, should we say something at the beginning of Section 4 to connect with the message of this Section? The two sections are a bit disconnected for now.}

To uncover the new factor causing forgetting, we deliberately consider the cases where the behavior of backbone w.r.t each learned task is unchanged over time - meaning (I) and (II) would not happen. Firstly, to eliminate concerns about changing learned parameters \textbf{(I)}, we consider methods that propose using a distinct set of prompts $\bm{P}_t$ to a specific task $t$. Thus, we conduct experiments on HiDE-Prompt \citep{wang2023hierarchical}, which is the latest SOTA in prompt-based CL. Then, the remaining potential factor is the difference between the prompt chosen at inference time. To this end, the training is carried out as usual; when testing, we intentionally choose the right prompt for each sample to ensure that there is no change in the backbone's behavior compared to training. And the features after backbone will be classified normally without taskID.

For a detailed examination, we propose an experimental setup to analyze the performance degradation of corresponding models over time. Specifically, we train a classification head $s_1(\bm{x})$ on the latent space $f_{\Phi, \bm{P}_{1}}(\bm{x})$ derived from the first task $\mathcal{D}_1$.
% to accurately classify the classes within this initial task. 
As the model undergoes continual training on a sequence of subsequent tasks, we assess the ability of $s_1(\bm{x})$ to classify instances from $\mathcal{D}_1$ over time,
% based on the latent space representation $f_{\Phi, \bm{P}_{\hat{t}}}(\bm{x})$, 
defining this as \textit{within-task accuracy}. This is different from \textit{model accuracy on} $\mathcal{D}_1$, which is obtained when we use the usual classification head $h_\psi$ as the original design of HiDE-Prompt. This classification head $h_\psi$ considers all classes that the model has observed so far, instead of just data of $\mathcal{D}_1$ as $s_1(\bm{x})$. The results in Figure \ref{fig:wt_a}, show that \textit{within-task accuracy} definitely remains. 
% suggesting minimal shift between representations $f_{\Phi, \bm{P}_{\hat{t}}}(\bm{x})$ and $f_{\Phi, \bm{P}_{t}}(\bm{x})$.
In contrast, Figure \ref{fig:true_a}, illustrates a significant decrease in \textit{true accuracy}, raising the question of whether we have overlooked additional factors contributing to final forgetfulness.

% ???? 
% To identify the cause of the performance degradation despite the negligible shift, we now investigate the \textit{inference feature space} of \( f_{\Phi, \bm{P}_{\hat{t}(\bm{x})}}(\bm{x}) \). 
% Although there is a minor shift between \( f_{\Phi, \bm{P}_{\hat{t}(\bm{x})}}(\bm{x}) \) and \( f_{\Phi, \bm{P}_{t(\bm{x})}}(\bm{x}) \), leading to good performance if the correct classification head is chosen, as more tasks arrive, the number of classes and classification heads increases. This results in greater overlap in the \textit{inference feature space}, causing more confusion for the classification heads when making predictions. 

\begin{figure}[ht]
    \centering
    \vspace{-8mm}
    \includegraphics[width=0.501\textwidth]{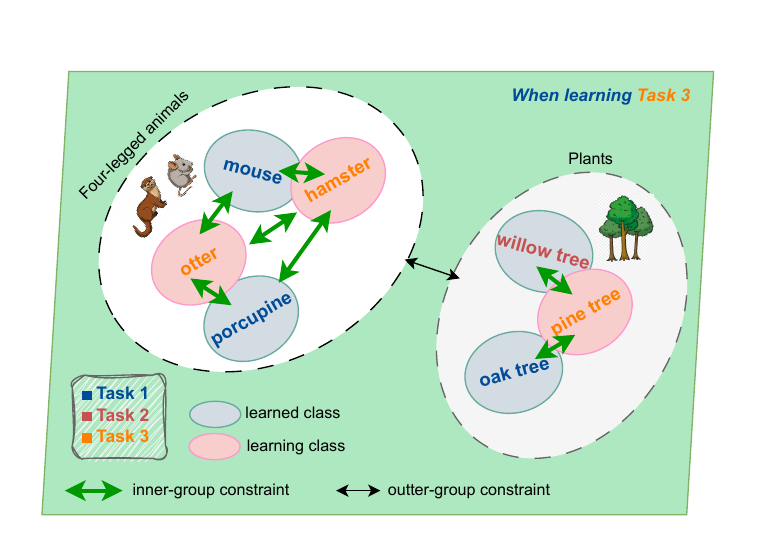}
    \vspace{-8mm}
    \caption{\textbf{\textit{Problem:}} When new classes arrive, the latent space of a model for all tasks so far becomes fuller, and class representations tend to be overlapped, leading to performance degradation in old tasks. \textbf{\textit{Our solution:}} We focus on separating easily confused classes whose concepts/labels lie in the same leaf group on a label-based hierarchical taxonomy, suggested by expert knowledge (i.e., ChatGPT - Figure \ref{fig:motivation_taxonomy}). \textbf{\textit{[Best viewed in color mode]}}}
    \label{fig:train}
    \vspace{-5mm}
\end{figure}

Additionally, considering the \textit{inference feature space} of \( f_{\Phi, \bm{P}}(\bm{x}) \), we can see that as more tasks arrive, the number of classes increase, making the space fuller and increasing the possibility of overlap between class distributions.
% To further demonstrate the increasing overlap and filling of the inference feature space, 
To support this point, we illustrate the representations of some class images in Figure \ref{fig:train}, as well as the corresponding t-SNE visualizations in
Figure \ref{fig:tnse}. In particular, after Task 1, we have representations of "oak tree", "mouse" and "porcupine" located in quite separate locations. However, when Task 2 and then Task 3 arrive, the appearance of "willow tree" and "pine tree" makes \textit{the latent space become fuller}, and "oak tree" no longer maintains the separation from the remaining classes as before, \textit{leading to a remarkable drop in performance of previous tasks}.  

Based on the evidence above, another key cause of catastrophic forgetting in prompt-based continual learning that should be recognized is \textit{the addition of new classes, which gradually fills the latent space and may overlap with existing ones}. The overlap causes confusion in distinguishing between old and new classes, thereby reducing the performance of learned tasks over time. This is coincidently similar to the behavior of human memory, where although the old knowledge exists somewhere in the brain, still be confused due to the newly acquired information \citep{anderson1996interference, wimber2015retrieval, loftus2005misinformation, nader2000reconsolidation, wixted2004forgetting}. While existing CL methods emphasize the importance of representation learning to keep classes distinct \citep{wang2023hierarchical, li2023steering}, none explicitly acknowledge the overlap between new and old tasks as a source of forgetting. 
Recognizing this motivates us to propose a novel method, focusing on identifying easily confused class pairs, thereby reducing forgetting and improving performance.

%% file: sec/3-framework.tex
In the previous section, we noted that increasing overlap in data representations as more tasks arrive is one of the main reasons leading to performance degradation (i.e., forgetting). Therefore, if we can identify easily confused classes/concepts, we will better enhance the distinguishability of models and thus reduce forgetting. Inspired by Cognitive Science studies, regarding the benefits of organizing information in a meaningful way (Section~\ref{method: motivation}),
we propose using expert knowledge to structure the concepts/labels of CL tasks in the corresponding hierarchical taxonomies. Interestingly, we find that concepts within the same group in each taxonomy tend to be visually and semantically similar, potentially causing more overlap in the visual latent space and causing confusion for the CL classifier. Motivated by this observation, we propose group-based contrastive learning in Section~\ref{taxooo} to maximize the separability of these concepts, thus mitigating forgetting over time. Furthermore, to reinforce the method's effectiveness, Section~\ref{ot_tricky} presents an optimal transport-based technique, harnessing priori of pretrained backbones from a new perspective.

\subsection{Motivation} 
\label{method: motivation}
% \paragraph{{Insights from Cognitive Science.}} Research has shown that humans better understand and remember information when it is organized or connected meaningfully \citep{bruner1956thinking, sweller1988cognitive, miller1956magical, rogers1977}. One approach that enhances memory retention is associative learning—the idea that information is easier to recall when linked with related concepts or prior knowledge. This type of learning involves strengthening the connections between neurons when they are activated together, a principle known as \textit{Hebbian learning}. Additionally, \textit{elaborative encoding} has been identified as a key technique for improving memory. Studies, such as those by \cite{rogers1977}, have found that when individuals process information in a more meaningful way, such as relating it to themselves or to familiar concepts, they are more likely to remember it. This method leverages the self-reference effect, where personalizing information during learning enhances recall.

\paragraph{Insights from Cognitive Science.} Research in Cognitive Science highlights the importance of \emph{reflection, and organization of information} as critical components for effective learning. Studies show that when learners take time to reflect on their experiences, they deepen their understanding and enhance retention \citep{schon1983reflective}. This reflective practice encourages individuals to connect new information with existing knowledge, fostering a more integrated learning experience \citep{bransford2000how}. Linking new material with relevant prior knowledge—often referred to as associative learning—further strengthens memory retention \citep{sweller1988cognitive} and also benefits future learning. Moreover, organizing information into coherent structures, such as outlines or concept maps, allows learners to see relationships between concepts, making it easier to retrieve the information later \citep{mayer2005cambridge}.

% \vspace{-3mm}
\begin{figure*}[ht]
    \centering
    \includegraphics[width=0.85\textwidth]{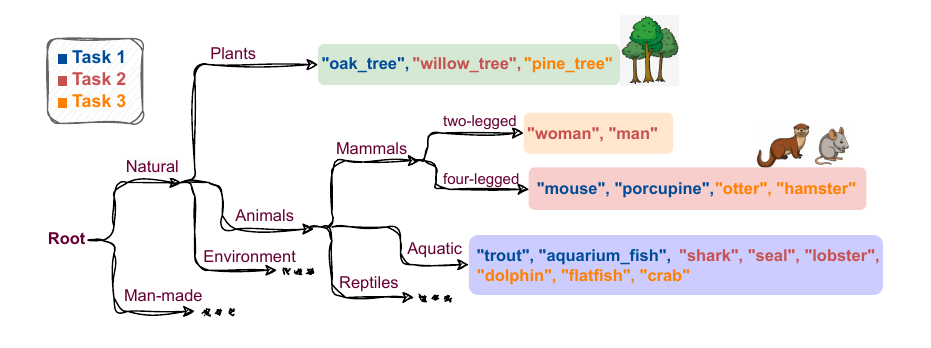}
    \vspace{-6mm}
    \caption{\textbf{The label-based hierarchical taxonomy} obtained when learning Task 3 on Split-CIFAR100. The colors (i.e., \textbf{\textcolor{blue}{blue}}, \textbf{\textcolor{purple}{red}}, \textbf{\textcolor{orange}{orange}}) of the label names represent the order in which the corresponding classes appear. Accordingly, the tree-like taxonomy is gradually developed and detailed over time. \textbf{\textit{[Best viewed in color mode]}} }
    \label{fig:motivation_taxonomy}
    % \vspace{-2mm}
\end{figure*}

\begin{figure*}[htbp]
    \centering
    % Subfigure 1
    \begin{subfigure}[b]{0.48\textwidth}
        \centering
        \includegraphics[width=0.9\textwidth, height=6.28cm]{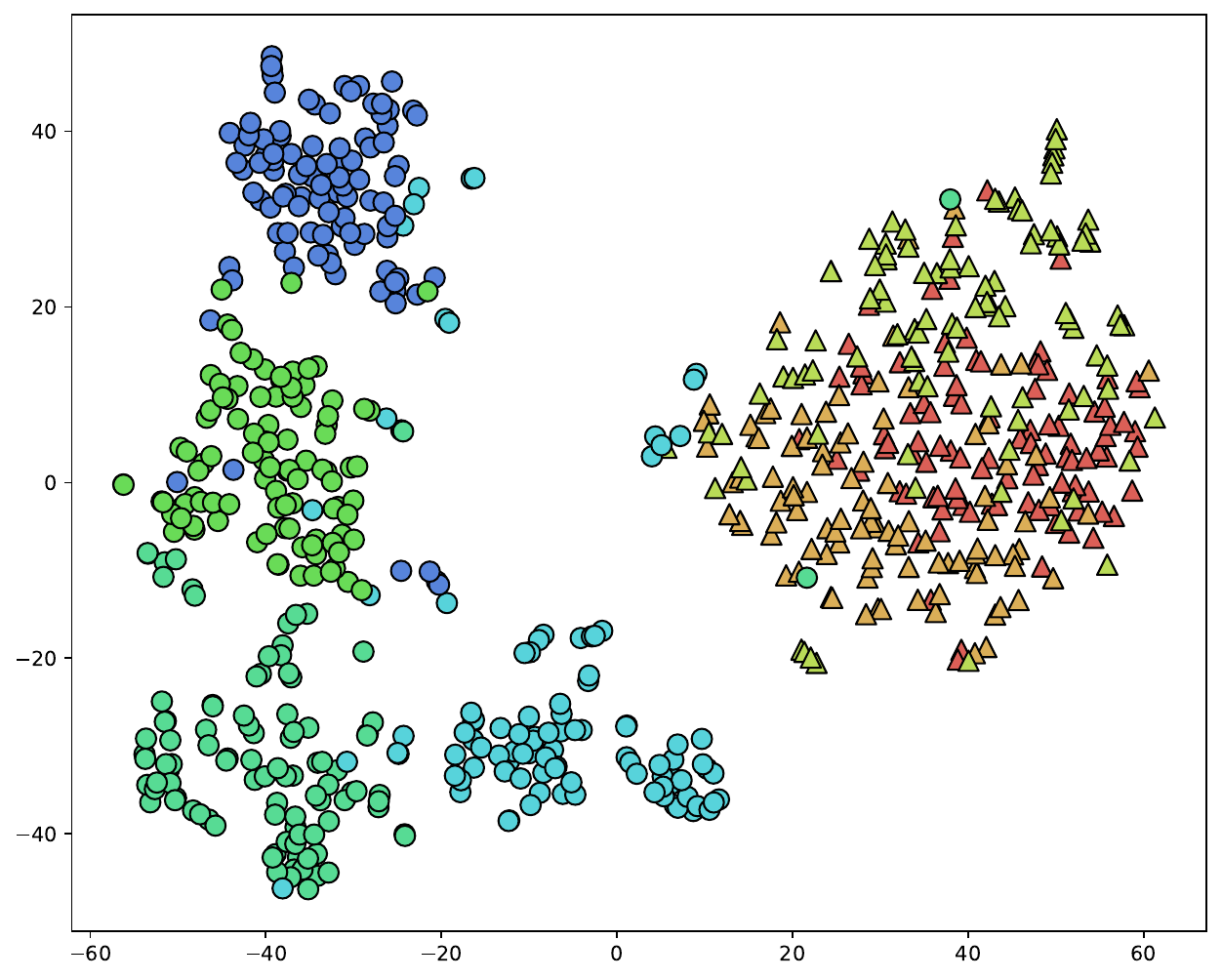} % Replace with your image file
        \caption{Training tasks independently with $\mathcal{L}_{CE}$}
        \label{fig:L_CE}
    \end{subfigure}
    \hfill % Adds horizontal space between the subfigures
    % Subfigure 2
    \begin{subfigure}[b]{0.48\textwidth}
        \centering
        \includegraphics[width=0.9\textwidth]{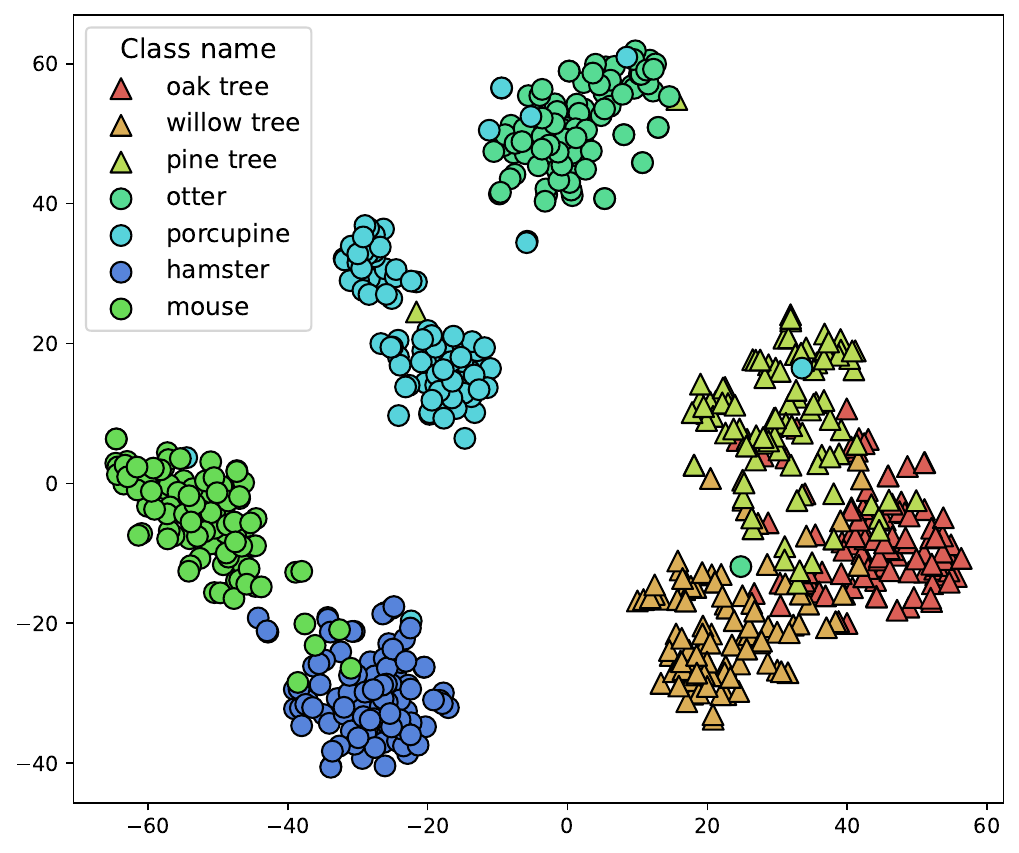} % Replace with your image file
        \caption{Ours (TCL)}
        \label{fig:L_G}
    \end{subfigure}
    \vspace{-2mm}
    \caption{\textbf{t-SNE visualizations} of classes within leaf groups of Four-legged animals ($ \bullet $ circular points) and Plants ($\blacktriangle$ triangular points) when learning Task 3, Split-CIFAR-100. The appearance order of the classes: Task 1 - \textit{"mouse", "porcupine", "oak tree";} Task 2 - \textit{"willow tree"}; Task 3 - \textit{"otter", "hamster", "pine tree"} (also refer to Figure \ref{fig:train} and \ref{fig:motivation_taxonomy}). We can see that if we train tasks independently with $\mathcal{L}_{CE}$, the classes within each leaf group, which arrive at different time, can be overlapped seriously (Figure \ref{fig:L_CE}).}
    \label{fig:tnse}
    \vspace{-3mm}
\end{figure*}

% Together, these strategies create a robust framework for learning, emphasizing that how information is processed and connected can significantly impact cognitive outcomes.
\vspace{-3mm}
\paragraph{Our Approach.} It is evident that
besides \emph{reflection} - comparison of old and new knowledge, 
the key factor in learning efficiently is \emph{to organize and link information in an insightful way}, where concepts are linked and arranged according to their semantic meanings. This observation motivates us to develop deep learning models that learn concepts structured in a hierarchical taxonomy. The aim is to \textit{enable the models to grasp relevant concepts more effectively}, thereby \textit{enhancing distinguishability and mitigating catastrophic forgetting}. More specifically, we propose \textit{structuring the data labels} in a tree-like taxonomy, which can be flexibly expanded over time using domain expertise. Based on this structure, we have a view of the relationships between classes, identifying which classes belong to the group with many shared characteristics, easily confused, and require more focus (see Figures \ref{fig:motivation_taxonomy}, \ref{fig:tnse}).

Taking the learning process of Split-CIFAR-100 (see Appendix \ref{Datasets}) as an example, when training on task \( t=3 \), we can construct a tree-like taxonomy of concepts/labels, as shown in Figure \ref{fig:motivation_taxonomy}. We observe that the classes under a same leaf in the structure (e.g., oak tree, willow tree, and pine tree in the leaf group "plants") exhibit stronger visual and semantic correlations. Consequently, as shown in Figure \ref{fig:tnse}a, these class features are more overlapping with each other rather than with another class from other leaf groups (e.g., "four-legged"). In this way, the tree-like taxonomy serves as a tool to help identify easily confused classes, facilitating the subsequent process of making them more separable in the feature space. 
Leveraging the insight obtained from the taxonomy above, we aim to train a network so that all class representations must be distinct, \textit{especially those within each leaf group}.
% For the leaf group of four-legged animals, we assume that "mouse" and "porcupine" arrive in Task 1, while "otter" and "hamster" are in Task 3.  When learning Task 1, to classify "mouse" and "porcupine", we encourage the model to capture the important features of four-legged mammals to efficiently differentiate between them.
% We hope that the knowledge learned from "mouse" and "otter" in Task 1 can be beneficial for the next tasks. 
% Then in Task 3, the model learns to distinguish "hamster", "otter" and the old ones in this group. In this way, the mechanism helps further strengthen the learning of more efficient and robust features for the Mammals group.
Over time, this strategy not only helps the model better mitigate forgetting caused by representation overlap but also reinforces existing knowledge to facilitate future learning.

% To summarize, by grouping and categorizing, we expect the model to concentrate more on the detailed features of each leaf group. This enhances its ability to distinguish related objects and reinforces the model's knowledge of each group. Therefore, this strategy not only alleviates the forgetting of old knowledge—often caused by new classes that are difficult to distinguish from old ones within the same leaf group, but also enables active knowledge transfer between tasks.

% \subsection{Training phase}
\subsection{Leveraging Hierarchical Taxonomies for Better Representation Learning}
\label{taxooo}

% \textbf{[Figure of method - how to develop the taxonomy over time, and how we strengthen knowledge about each group...]}

During the training process, whenever a new class appears, its label name is automatically added to the tree-like taxonomy, into a leaf group containing classes with similar characteristics (Figure \ref{fig:motivation_taxonomy}). To develop this hierarchical structure, we can rely on expert knowledge, for example, ChatGPT, which can help incrementally construct a meaningful related tree (see Appendix \ref{chatGPT}). Structuring information in this way not only aligns with how the human brain effectively connects and remembers information but also \textit{provides useful insights during training, indicating how each knowledge is related to the other and which requires further focus.}

% \begin{wrapfigure}{r}{0.55\textwidth}
%     \centering
%     % \vspace{-10mm}
%     \includegraphics[width=0.6\textwidth]{img/train.pdf}
%     % \vspace{-4mm}
%     \caption{We focus on separate easily confused classes within each leaf group.}
%     \label{fig:train}
% \end{wrapfigure}

% As analyzed above, reflecting on and organizing knowledge are the key factors for efficient learning. That is, \textit{the model should always be encouraged to identify the decision boundary between all old and new classes, especially those with common characteristics}. 
Particularly, for each task $t$, we dedicate a set of prompt $\bm{P}_t$. We aim to minimize overlap between all classes so far, especially focusing on increasing the separability between classes belonging to the same leaf group on the taxonomy (e.g., four-legged mammals, plants, etc.,). That is, \textit{taxonomy acts as a reference information channel to support the training process.} Let $g$ be a leaf group on the taxonomy $\mathcal{G}$ (i.e., $g \in \mathcal{G}$), ${X}^g_{k}$ and ${Y}^g_{k}$  denote the corresponding sets of input samples and labels under the group $g$ that belong to the task $k$ ($k \leq t$). Besides Cross Entropy loss $\mathcal{L}_{CE}$, we propose using a regularization loss function for sample $\vx$ that arrives in task $t$ and belongs to leaf group $g$ as follows:
\begin{equation}
    \mathcal{L}_\mathcal{G}(\psi, \bm{P}_t, \bm{x}) = \alpha \mathcal{L}_{g}(\psi, \bm{P}_t, \bm{x}) + \beta \mathcal{L}_{all}(\psi, \bm{P}_t, \bm{x})
    \label{group_loss}
\end{equation}
where we have defined
\begin{align*}
    \mathcal{L}_{g}(\cdot) &= -\log \sum_{\vx'\in X^g_{1...t}|y_{\vx'}=y_\vx} \frac{u(\vz_x \cdot  \vz_{x'})}{\sum_{\bar{x} \in X^g_{1...t}} u(\vz_x \cdot  \vz_{\bar{x}}) 
    } \\
    \displaystyle \mathcal{L}_{all}(\cdot) &= -\log \sum_{\vx'\in X_{1...t}|y_{\vx'}=y_\vx} \frac{u(\vz_x \cdot  \vz_{x'})}{\sum_{\bar{x} \in X_{1...t}} u(\vz_x \cdot  \vz_{\bar{x}})
    }
\end{align*}
are the Supervised Contrastive losses that we put on class representation \textit{within leaf group} $g$, and \textit{all classes} so far, respectively; and
\begin{equation}
    u(\vz_\vx \cdot  \vz_{\vx'}) = \exp( \frac{\vz_\vx \cdot  \vz_{\vx'}}{\tau})
    \label{uuuu}
\end{equation}
% u(\vz_\vx \cdot  \vz_{\vx'}) = \exp( \frac{\vz_\vx \cdot  \vz_{\vx'}}{\tau})$,
with $\vz_\vx = f_{\Phi, \bm{P}_t}(\vx)$ is the feature vector on the latent space of the prompt-based model and $y_\vx$ is the ground truth label of $\vx$, $\tau$ is the temperature with $\tau =0.1$ for all experiments, and $\alpha$ is the coefficient that controls how much we want to force classes within leaf group $g$ stay apart further.
% \color{red}Tuan: maybe explain the hyperparameter $\alpha$ here, I would say something like "the hyperparameter $\alpha$ controls how much we want to enforce the representation $\vz_x$ to be close to its corresnponding leaf group with the same label"\color{black}. 
In addition, because the data of old tasks can not be accessed when training the new one, we follow previous work \citep{wang2023hierarchical, li2024steering} to encode the information of each trained class $c$ in the form of a Gaussian Mixture model $\text{GMM}_{c} = \{ \mathcal{N}(\mu_{c, i}, \Sigma_{c, i}) \}_{i=1}^{K}$, at the end of the corresponding task, where $K$ is the number of components. Then in task $t$, the representation of each old class $c$ is sampled from $\text{GMM}_{c}$, to compute loss functions in Eq.~(\ref{group_loss}).
% , supporting the training of the current task. 
% The technique of using pseudo features of old data is also employed in many existing prompt-based CL methods, as the shift of features is minimal (Table \ref{tab:shift}). 

Looking closer at Eq.~(\ref{group_loss}), this equation implies that when learning a new task, the model will be encouraged to identify the decision boundary between all old and new classes (i.e., using $\mathcal{L}_{all}$). Besides, it will especially focusing on those in the same leaf group, which often share many common characteristics and can confuse classification models (i.e., using $\mathcal{L}_{g}$), see Figure \ref{fig:train}. 
This manipulation matches the way our brain naturally works - without reflection and comparison, we would be confused about old and new concepts, especially those sharing many common characteristics, leading to incorrect judgments and decisions in practice. For AI models, learning new tasks without thoroughly considering the learned information of old tasks can lead to uncontrolled overlapping in the latent space, resulting in forgetting the learned knowledge and harming final performance. In this work, thanks to the insight from the hierarchical taxonomy, we can identify which group of classes that easy to get overlapped in the latent space, thereby actively intervening and enhancing the model representation learning in the CL environment.

\subsection{Harnessing Prior Knowledge from Pretrained Models via Optimal Transport}
\label{ot_tricky}

% Although the taxonomy constructed above is a useful information channel that provides a new perspective on the relationship between data, the tree structure is still fundamentally based on the semantic information of the text labels. Condensing information into a few labels is sometimes not enough to describe the myriad characteristics of visual data. 

% \begin{wrapfigure}{r}{0.5\textwidth}
%     \centering
%     \vspace{-5mm}
%     % \includegraphics[width=0.55\textwidth]{img/WS_latent_space.pdf}
%     \includegraphics[width=0.5\textwidth]{img/WS_latent_space_bblueue.pdf}
%     \vspace{-4mm}
%     \caption{Wasserterin distance between classes (Split-CIFAR100) in latent space of pretrained backbone (Sup-21K).}
%     % \vspace{-3mm}
%     \label{fig:ot_trick}
% \end{wrapfigure}

\begin{figure}[ht]
    \centering
    \includegraphics[width=0.8\linewidth]{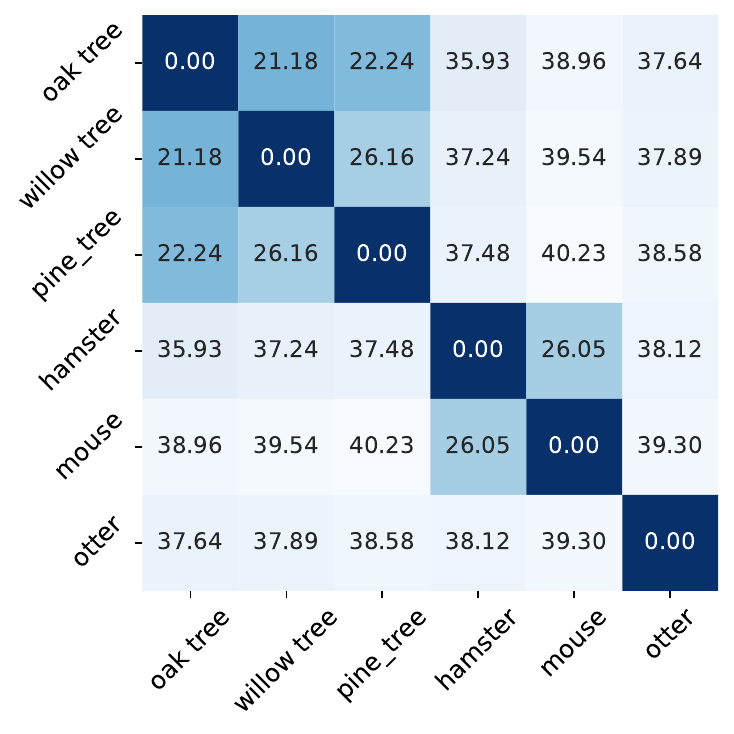}
    \vspace{-4.5
    mm}
    \caption{L2-Wassertein distance between classes (Split-CIFAR-100) in latent space of a pretrained backbone (Sup-21K). \textbf{\textit{Within a leaf group}} (i.e., "plant" or "four-legged mammals"), \textit{\textbf{there may be data classes with varying levels of correlation}} in the latent space.}
    \label{fig:ot_trick}
    \vspace{-5mm}
\end{figure}

Considering a leaf group, there may be data classes with varying levels of overlap in the latent space. For example, Figure \ref{fig:ot_trick} provides statistical results regarding the relative position of classes in leaf groups "plants" and "four-legged mammals" in the latent space of a pretrained model. We can see that in the group of animals, the L2-Wasserstein distance between class representations of "mouse" and "otter" (39.30) is more significant than the one between "mouse" and "hamster" (26.05),  meaning that the class representation of the second pair can overlap more than the first one.
This empirical result suggests that although the strategy in Sec \ref{taxooo} helps improve the model's ability to recognize difficult-to-identify classes, we still treat all classes in that group equally. Thus, \textit{the algorithm may inadvertently ignore the pairs of classes that are easily confused and need to be further distinguished.}

Moreover, pretrained models are extensively trained on large datasets, which results in their substantial generalization abilities. That is the reason why they are considered as good starting points for the adaptation to downstream tasks \citep{devlin2019bert, brown2020gpt3, he2016resnet, raffel2020t5}. 
% However, the existing methods seem to overlook the initial behavior of pretrained models on the training data, particularly regarding relations between classes - which classes are easily classified and which are prone to confusion. 
In this work, we propose utilizing the pretrained models from another novel perspective, which can comprehend the use of the label-based taxonomy (Sec. \ref{taxooo}) during training. Particularly, we take advantage of these models to obtain prior assumptions about the relationships between the class representations, thereby elegantly intervening in better separating each pair of classes within each leaf group.  

\begin{table*}[!ht]
	\centering
    \vspace{-0.37cm}
    
    % \vspace{-0.3cm}
    %\cite{smith2022coda} uses self-supervised pretraining on ImageNet-21K with supervised fine-tuning on ImageNet-1K, and different data splits for ImageNet-R. Here we reproduce its results with the same settings.
    
      %\vspace{-0.1cm}
	\smallskip
      \renewcommand\arraystretch{1.5}
    \small{
	\resizebox{1.0\textwidth}{!}{ 
	\begin{tabular}{l cc cc cc cc}
	 \hline
        % \multirow{2}{*}{PTM} & \multirow{2}{*}{\,\,\,\,\,\,\,\,\,\,\,\,\,\,\,\,\,\,\,\,\,\,\,\, Method} & \multicolumn{3}{c|}{Split CIFAR-100} & \multicolumn{3}{c|}{Split ImageNet-R} & \multicolumn{3}{c|}{5-Datasets} & \multicolumn{3}{c}{Split CUB-200} \\
        % & & \textbf{FAA} ($\uparrow$) & CAA ($\uparrow$) & FFM ($\downarrow$) & \textbf{FAA} ($\uparrow$) & CAA ($\uparrow$) & FFM ($\downarrow$) & \textbf{FAA} ($\uparrow$) & CAA ($\uparrow$) & FFM ($\downarrow$) & \textbf{FAA} ($\uparrow$) & CAA ($\uparrow$) & FFM ($\downarrow$)\\
        \multirow{2}{*}{Method} & \multicolumn{2}{c}{Split CIFAR-100} & \multicolumn{2}{c}{Split ImageNet-R} & \multicolumn{2}{c}{5-Datasets} & \multicolumn{2}{c}{Split CUB-200} \\
        \cmidrule(lr){2-3} \cmidrule(lr){4-5} \cmidrule(lr){6-7} \cmidrule(lr){8-9}
        & \textbf{FAA} ($\uparrow$)  & FFM ($\downarrow$) & \textbf{FAA} ($\uparrow$) &  FFM ($\downarrow$) & \textbf{FAA} ($\uparrow$)  & FFM ($\downarrow$) & \textbf{FAA} ($\uparrow$) &  FFM ($\downarrow$)\\
        \hline
    %    \multirow{6}*{\tabincell{c}{Sup-21K}} 
       L2P &83.06 \small{$\pm 0.17$} &6.58 \small{$\pm 0.40$} &63.65 \small{$\pm 0.12$} &7.51 \small{$\pm 0.17$} & 81.84 \small{$\pm 0.95$}  & 4.58 \small{$\pm 0.53$} & 74.52 \small{$\pm 0.92$} & 11.25 \small{$\pm 0.23$} \\ 
       
       DualPrompt &86.60 \small{$\pm 0.19$} &4.45 \small{$\pm 0.16$} &68.79 \small{$\pm 0.31$} &4.49 \small{$\pm 0.14$} &77.91\small{$\pm 0.45$} & 13.17 \small{$\pm 0.71$} & 82.05\small{$\pm 0.95$} & 3.56 \small{$\pm 0.53$} \\ 

       OVOR & 86.68 \small{$\pm 0.22$} & 5.25 \small{$\pm 0.12$} & \underline{75.61 \small{$\pm 0.82$}}& 5.77 \small{$\pm 0.12$} & 82.34 \small{$\pm 0.48$} & 4.83 \small{$\pm 0.35$} & 78.12 \small{$\pm 0.0.65$} & 8.13 \small{$\pm 0.52$} \\ 
       \hline 
       S-Prompt++ &88.81 \small{$\pm 0.18$} &3.87 \small{$\pm 0.05$} & 69.68 \small{$\pm 0.12$} &4.25 \small{$\pm 0.05$} &86.19\small{$\pm 0.65$} & 4.67 \small{$\pm 0.72$} & 83.12 \small{$\pm 0.54$} & 2.72 \small{$\pm 0.64$} \\
       
       CODA-Prompt &86.94 \small{$\pm 0.63$} &4.04 \small{$\pm 0.18$} & 70.03 \small{$\pm 0.47$}& 5.17 \small{$\pm 0.22$} & 64.20 \small{$\pm 0.53$} & 17.22 \small{$\pm 0.55$} & 74.34 \small{$\pm 0.68$}  & 12.05 \small{$\pm 0.41$} \\

       CPP &91.12 \small{$\pm 0.12$} & 3.33 \small{$\pm 0.18$} & 74.88 \small{$\pm 0.07$} & 4.08 \small{$\pm 0.03$} & 92.92 \small{$\pm 0.17$} & \underline{0.23 \small{$\pm0.07$}} & 82.35 \small{$\pm0.23$} & 3.24 \small{$\pm0.32$} \\
       
       HiDe-Prompt  &\underline{{92.61} \small{$\pm 0.28$}} &\underline{{1.52} \small{$\pm 0.10$}}  &{75.06 \small{$\pm 0.12$}} & \underline{4.05 \small{$\pm 0.19$}} & \underline{93.92 \small{$\pm 0.33$}}  & 0.31 \small{$\pm 0.12$}  & \underline{86.62 \small{$\pm 0.35$}} & \underline{2.55 \small{$\pm0.15$}} \\
       
       \midrule
    
        % Ours (train only) & \textbf{93.94} \small{$\pm 0.22$} &  \textbf{1.56} \small{$\pm 0.18$} & \textbf{76.41} \small{$\pm 0.12$} & \underline{3.12 \small{$\pm 0.25$}} & \textbf{95.02} \small{$\pm 0.18$} & \underline{0.25 \small{$\pm 0.16$}} & \textbf{87.93} \small{$\pm 0.18$} & \underline{2.52 \small{$\pm 0.22$}} \\

        Ours (TCL) & \textbf{93.94} \small{$\pm 0.23$} &  \textbf{1.05 \small{$\pm 0.15$}} & \textbf{76.41} \small{$\pm 0.12$} & \textbf{4.02 \small{$\pm 0.22$}} & \textbf{95.02} \small{$\pm 0.20$} & \textbf{0.21 \small{$\pm 0.15$}} & \textbf{87.93} \small{$\pm 0.22$} & \textbf{2.01 \small{$\pm 0.23$}} \\
       
       \midrule
       
       % \hline
       
	\end{tabular}
	} }
        \vspace{-2mm}
        \caption{\textbf{Overall performance comparison.} We provide FAA and FFM of all methods, with standard deviation taken over at least 3 runs of different random seeds. The results corresponding to the best FAA among baselines are underlined.}
	\label{table:main_results}
	\vspace{-0.3cm}
\end{table*}

More specifically, to extract the relationship between the classes, we compute the L2-Wasserstein distance of feature distributions of each class pair. We denote $D_{c_i}^\Phi$ be the distribution of class $c_i$ on the latent space of the pretrained model $f_\Phi$ - to distinguish it from the one on prompted latent space $f_{\Phi, \bm{P}}$, as described in Sec. \ref{taxooo}, \textit{Line 364}.
The distribution $D_{c_i}^\Phi$ is obtained in the form of a Gaussian Mixture model at the end of the respective task, \textit{only once and then kept fixed for the next reuse; therefore, it does not impose a computational burden}. 
In this way, when $t$ tasks have arrived, we have the corresponding sets of distributions $\{ D_c^\Phi \}_{c \in Y_{1,t}}$ of all $m_t$ classes so far. Thus, we can gradually complete the WD-based matrix between pairs of classes:
\begin{equation}
    M= [W_2(D_{c_i}^\Phi, D_{c_j}^\Phi)]_{m_t\times m_t}.
\end{equation}

The fact is that the smaller the $W_2$ between the representations of two classes, the closer they are, and the more force can be needed to separate them. Thus, we employ these results to obtain the corresponding weight factors, which control how much constraint needs to be applied to each class pair. Particularly, we compute \textit{the weight matrix}:
\begin{equation}
    \displaystyle \Gamma = \left[ \textcolor{blue}{\bm{\gamma_{ij}}}
\right]_{m_t \times m_t} = \left[ 1/ {\exp(M_{ij}/\delta)} \right]_{m_t\times m_t}
\label{deltaaa}
\end{equation} where $\delta$ is a temperature. We then apply this information to obtain a weighted version of $\mathcal{L}_g$, in which the closer the two class distributions are, the larger the weight assigned, and they will be further focused to push away: 
% Consequently, our regularization loss becomes:
\begin{equation}
    \mathcal{L}^{'}_g(\cdot) = -\log \sum_{\vx'\in X^g_{1..t}|y_{\vx'}=y_\vx} \frac{u(\vz_x \cdot  \vz_{x'})}{\sum_{\bar{x} \in X^g_{1..t}} \textcolor{blue}{\bm{\gamma_{y_xy_{\Tilde{x}}}}}u(\vz_x \cdot  \vz_{\bar{x}}) 
    } 
    \label{group_loss_gamma}
\end{equation} 
This strategy is completely economical and aligns well with the CL learning scheme, as the distance between representations of each class pair only needs to be computed once, and the matrix $M$ is continuously expanded when new classes arrive.
Practically, when learning a new task, the first epoch is for capturing information about the behavior of the pretrained model on the data of this task. Moreover, this approach is also aligned with the findings in Cognitive Science \citep{osgood1953communication, baltes1987experience}, showing that the accumulated experiences from past learning create momentum for future learning.

Finally, the final objective function of our full method can be formulated as follows: 
% \color{red}
\begin{equation}
    \mathcal{L} = \mathcal{L}_{CE} + \alpha \mathcal{L}^{'}_g + \beta \mathcal{L}_{all}
    \label{final_loss}
\end{equation}

%% file: sec/4-exp.tex
\subsection{Experimental setup}

\textbf{Datasets.} We use 4 common CIL benchmarks, including Split CIFAR-100, Split ImageNet-R, 5-Datasets, and Split CUB-200 (see Appendix \ref{Datasets}).

\textbf{Baselines.} We compare ours with the various state-of-the-art prompt-based methods, including 
% the methods using shared prompts for all tasks: 
L2P \citep{wang2022learning_l2p}, DualPrompt \citep{wang2022dualprompt}, OVOR \citep{DBLP:conf/iclr/HuangCH24},
% ; and the methods dedicates a distinct prompt set for each task like: 
S-Prompt++ \citep{wang2022sprompts}, CODA-Prompt \citep{smith2022coda}, HiDe-Prompt \citep{wang2023hierarchical}, CPP \citep{li2024steering} (see Appendix \ref{baselines}). 

\textbf{Metrics.} We use two main metrics, including the Final Average Accuracy (FAA) and the Final Forgetting Measure (FFM)
% denoting the average accuracy and forgetting of all tasks after learning the sequence of tasks, respectively 
(see Appendix \ref{baselines}).
Besides, the detailed implementation is described in Appendix \ref{Implementation}.

% For convenience, we denote two variations of our method without the self-improvement process: (i) \textit{\textbf{BoostCL-s}} is the version that uses the task-adaptive backbone with a single classifier for the CP-based backbone, while (ii) \textit{\textbf{BoostCL-m}} is the version that utilizes our proposed ensemble classifier with the boosting strategy for multiple views.

% \vspace{-5mm}

\begin{figure*}[htbp]
    \centering
    % Subfigure 1
    \begin{subfigure}[b]{0.33\textwidth}
        \centering
        \includegraphics[width=\textwidth]{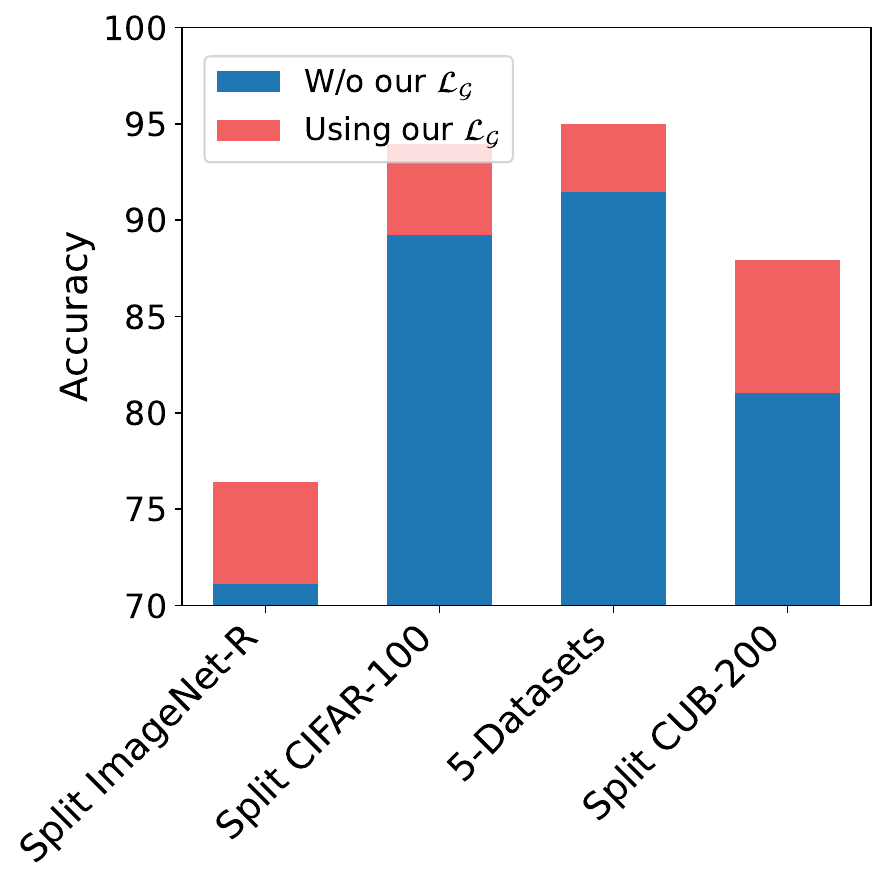} % Replace with your image file
        \caption{The role of our $\mathcal{L_G}$ - Eq (\ref{group_loss})}
        \label{fig:ablate_LG}
    \end{subfigure}
    \hfill
    % Subfigure 2
    \begin{subfigure}[b]{0.33\textwidth}
        \centering
        \includegraphics[width=\textwidth]{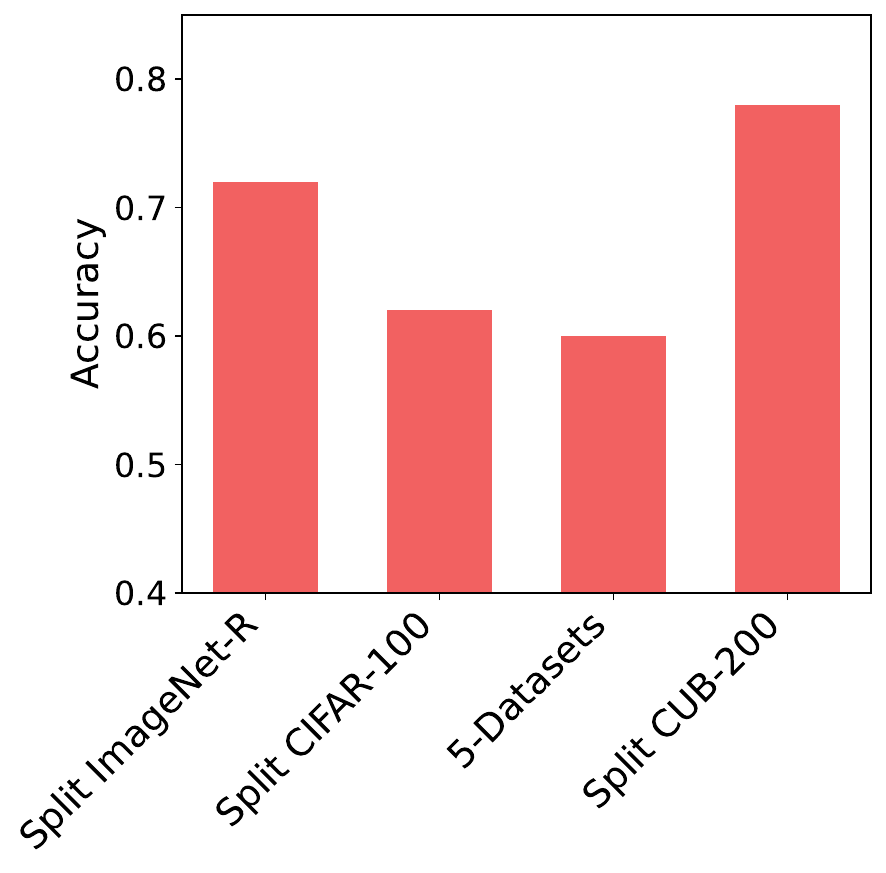} % Replace with your image file
        \caption{The role of our OT technique (Sec \ref{ot_tricky}) in $\mathcal{L_G}$}
        \label{fig:ablateOT}
    \end{subfigure}
    \hfill
    % Subfigure 3
    \begin{subfigure}[b]{0.325\textwidth}
        \centering
        \includegraphics[width=0.95\textwidth]{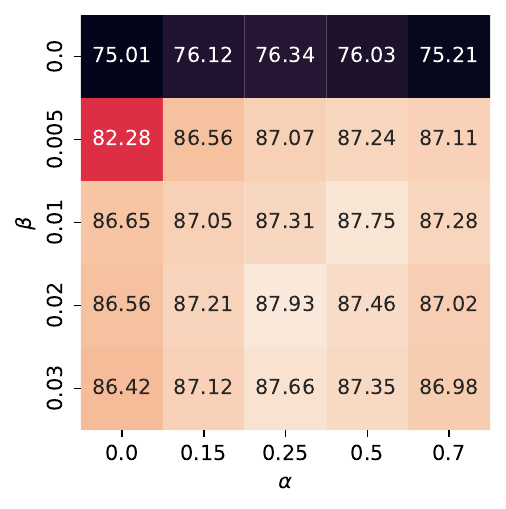} % Replace with your image file
        \vspace{8mm}
        \caption{Varying the value of $\alpha, \beta$ (Split-CUB-200)}
        \label{fig:ablate_alphabeta}
    \end{subfigure}
    \vspace{-2mm}
    \caption{Ablation studies about our training strategy. Figure \textbf{\textit{(b)}} depicts the performance improvement after applying our OT-based technique to our taxonomy strategy. The efficiency of this technique \textit{w/o our taxonomy strategy} is presented in Appendix \ref{ot_onlyy}.}
    \label{fig:ablate_train}
    \vspace{-3mm}
\end{figure*}

\subsection{Experimental result}

\paragraph{Our approach achieves superior results compared to baselines.} Table \ref{table:main_results} presents the overall performance comparison between our proposed method and other baselines. The key observation is that our method is the strongest one, because the gap compared with the runner-up method being around 1.5\% of FAA on all the datasets. Additionally, our method avoids forgetting better than all baselines, by the gap up to 17\% on 5-Datasets.
\vspace{-3.5mm}
\paragraph{Ablation studies.} 
Figure \ref{fig:ablate_train} reports the ablation studies of our training strategy. Particularly, compared to training tasks independently using Cross Entropy loss $\mathcal{L}_{CE}$ like in DualP and L2P, exploiting the relationships between data with the label-based taxonomy and the OT-based strategy (ours) helps improve FAA by about 5\% to 10\% (Figure \ref{fig:ablate_LG}). Besides, when examining the role of exploiting additional prior information from pretrained backbones using the OT approach upon the taxonomy, we see that FAA is improved from 0.6\% to 0.8\% (Figure \ref{fig:ablateOT}). These results demonstrate the positive impact of this component. In both figures, the improvements on Split-CIFAR100 and 5-Datasets are the lowest, while it is more pronounced on Split-CUB-200. This may be because the groups of these two datasets (Split-CIFAR100 and 5-Datasets) have fewer overlapping classes, as the classes in each group likely have more recognizable features. Meanwhile, Split-CUB-200 is a dataset of birds with images that can be difficult for human eyes to recognize, thus so our method performs better.
Furthermore, Figure \ref{fig:ablate_alphabeta} provides the experimental results when varying $\alpha$ and $\beta$.
% , which control the intensification of constraints on each leaf group of $\mathcal{L_G}$ during training. 
The data show significant degradations of model performance when each loss function of $\mathcal{L_G}$ is eliminated, demonstrating their specific roles. In particular, w/o using our taxonomy and OT-based strategy ($\alpha \mathcal{L}_g=0$), the performance can decrease up to 5.65\%. Besides, the effect of $\mathcal{L}_{all}$ can be clearly seen when $\beta=0$, as it helps constrain the relative correlations of all classes.
% The data show that with a large enough value of $\beta$, our TCL is not sensitive to $\alpha$ within its acceptable range. Conversely, if $\alpha$ is small, the quality of the model can change more significantly.

\vspace{-4.4mm}
\paragraph{t-SNE visualization on latent space.}Figure \ref{fig:tnse} illustrates the effect of our method in improving the model's representation learning on Split-CIFAR100. Specifically, the classes are better clustered, and the separation between them is more distinct. Especially, the classes "oak tree", "willow tree", and "pine tree" are divided into clear clusters, rather than being mixed together as in the traditional training strategy, where tasks are trained independently.
\vspace{-4mm}

\paragraph{The impact of $\delta$ in OT technique.}

\begin{figure}[!ht]
    \centering
    \vspace{-3mm}
    \includegraphics[width=1.\linewidth]{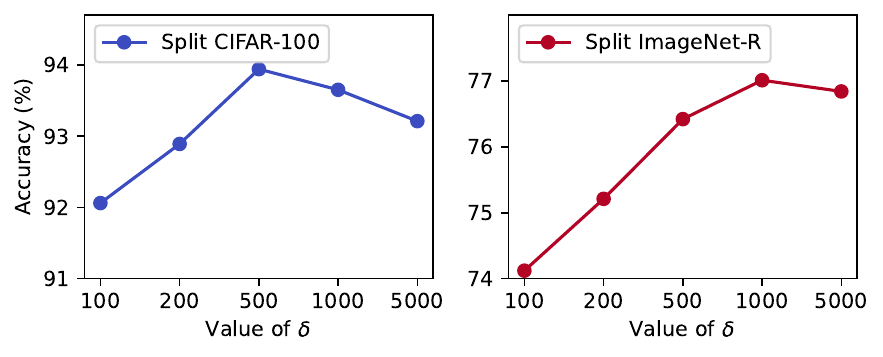}
    \vspace{-8mm}
    \caption{Model performance when $\delta$, in Eq (\ref{deltaaa}), varies.}
    \label{fig:delta_vary}
    \vspace{-3mm}
\end{figure}

Figure \ref{fig:delta_vary} illustrates the dependence of model performance on the value of the coefficient $\delta$ in our OT technique (see Eq.\ref{deltaaa}). The results show that if $\delta$ is too small, the performance can go down significantly, even worse than HiDE-Prompt. This occurs because the values of \textit{the weight matrix} will be too large, leading to classes, which have strong correlations in the latent space of $f_{\Phi}$, being pushed too far apart. This causes an imbalance compared to the overall correlation of all classes and unexpected overlapping. Conversely, if $\delta$ is set too large, the corresponding weight will be small, preventing achieving optimal efficiency. For Split-CIFAR-100 and Split-ImageNet-R, the optimal values are $\delta = 500$ and $100$, respectively.

\vspace{-7mm}
\paragraph{The superiority of our proposed method on various types of pretrained backbones.}Table \ref{table:backbone} illustrates that our method with the training strategy only consistently outperforms the strongest baseline (HiDE-Prompt) by the gap from about 0.5\% to 1.5\% in all cases.

\begin{table}[ht]
    \vspace{-0.2cm}
    \centering
    
     % \vspace{-0.2cm}
	\smallskip
      \renewcommand\arraystretch{1.2}
    \small{
	\resizebox{0.45 \textwidth}{!}{ 
        \centering
	\begin{tabular}{lcccc}
	 \hline
      \multirow{2}{*}{Backbone} & \multicolumn{2}{c}{Split CIFAR-100} & \multicolumn{2}{c}{Split Imagenet-R} \\
      \cmidrule(lr){2-3} \cmidrule(lr){4-5} 
        & TCL & HiDE-Prompt & TCL & HiDE-Prompt \\
        \hline
       Sup-21K & \textbf{93.94} & 92.61 & \textbf{76.41} & 75.06\\
       iBOT-21K & \textbf{94.01} & 93.02 & \textbf{72.12} & 70.83\\
       iBOT-1K & \textbf{94.27} & 93.48 & \textbf{72.80} & 71.33\\
       DINO-1K & \textbf{94.12} & 93.51 & \textbf{69.25} & 68.11\\ 
       MoCo-1K & \textbf{92.32} & 91.57 & \textbf{64.23} & 63.77 \\
       \hline
	\end{tabular}
	} 
}
        \vspace{-2mm}
        \caption{Comparison when using different pretrained backbones.} 
	\label{table:backbone}
	\vspace{-0.2cm}
\end{table}
\vspace{-6mm}
\paragraph{Other results.} We also provide experimental results in Appendix sections, related to model performance when using different taxonomy structures, and analyze other aspects of our OT technique, the coefficients $\alpha, \beta, \tau$ when using loss functions. Additionally, we verify the correlation between taxonomy and visual space, as well as discuss the current limitations and potential impacts of the proposed method. See Appendix \ref{additional_exp}.

%% file: sec/5-conclusion.tex
In this work, we demonstrate the importance of meaningfully organizing data information rather than just lumping them together for training. Particularly, we propose arranging data labels into tree-like taxonomies to identify data groups that are likely to confuse models.  This approach encourages the models to focus and develop deeper knowledge about each group, reducing forgetting and motivating more effective learning in subsequent tasks. Additionally, we propose leveraging the initial behavior of pretrained models to obtain hidden structures of training data, providing a new perspective to further enhance performance. Finally, the experimental results demonstrate our superiority over state-of-the-art baselines.

% \newpage
% \paragraph{Limitations.}Despite this novel approach, the quality of the hierarchical taxonomy depends on the quality of expert knowledge. For example, if similar image classes are not assigned to the same leaf group in this label-based taxonomy, the constraint we put on each such group may not perform as expected. 

%% file: sec/6-appendix.tex
% \onecolumn
% \maketitlesupplementary

% \maketitle

\clearpage
\setcounter{page}{1}
\maketitlesupplementary
% \LARGE Appendices 
% \progtitel{Appendices}
% \begin{center}
% \textbf{\Large{Supplement to
% ``Leveraging Hierarchical Taxonomies in Prompt-based Continual Learning``}}
% \end{center}

% \maketitle

\section{Experimental Settings}
\subsection{Datasets}
\label{Datasets}
We adopt the following common benchmarks:
\begin{itemize}
    \item \textbf{Split CIFAR-100} \citep{krizhevsky2009learning_cifar}: This dataset includes images from 100 different classes, each being relatively small in size. The classes are randomly organized into 10 sequential tasks, with each task containing a unique set of classes.
    
    \item \textbf{Split ImageNet-R} \citep{krizhevsky2009learning_cifar}: This dataset contains images from 200 extensive classes. It includes difficult examples from the original\textbf{ ImageNet} dataset, as well as newly acquired images that display a variety of styles. The classes are randomly divided into 10 distinct incremental tasks.
    
    \item \textbf{5-Datasets} \citep{ebrahimi2020adversarial}: This composite dataset incorporates \textbf{CIFAR-10} \citep{krizhevsky2009learning_cifar}, \textbf{MNIST} \citep{lecun1998gradient_mnist}, \textbf{Fashion-MNIST} \citep{xiao2017fashion}, \textbf{SVHN} \citep{netzer2011reading}, and \textbf{notMNIST} \citep{bulatov2011notmnist}. Each of these is treated as a separate incremental task, enabling the evaluation of the impact of substantial variations between tasks.
    
    \item \textbf{Split CUB-200} \citep{wah2011caltech}: This dataset contains fine-grained images of 200 distinct bird species. It is randomly divided into 10 incremental tasks, each with a unique subset of classes.

\end{itemize}

\subsection{Baselines}
\label{baselines}
In the main paper, we use CL methods with pretrained ViT as the backbone. We group them into (a) the group using a common prompt pool for all tasks, and (b) the group dedicating distinct prompt sets for each task:

(1) \textbf{L2P} \citep{wang2022learning_l2p}: The first prompt-based work for continual learning (CL) suggested using a common prompt pool, selecting the top $k$ most suitable prompts for each sample during training and testing. This approach might facilitate knowledge transfer between tasks but also risks catastrophic forgetting. Unlike our approach, L2P doesn't focus on training classifiers or setting constraints on features from old and new tasks during training, which may limit the model's predictability.

% This approach not only helps pretrained ViT efficiently adapt to new downstream tasks but also allows for the ability to transfer knowledge between tasks.

(2) \textbf{DualPrompt} \citep{wang2022dualprompt}: The prompt-based method aims to address L2P's limitations by attaching complementary prompts to the pretrained backbone, rather than only at input. DualP introduces additional prompt sets for each task to leverage task-specific instructions alongside invariant information from the common pool. However, like L2P, it does not focus on efficiently learning the classification head. Additionally, selecting the wrong prompt ID for task-specific instructions during testing can negatively impact model performance.

(3) \textbf{OVOR} \citep{DBLP:conf/iclr/HuangCH24}: while using only a common prompt pool for all tasks, this work introduces a regularization method for Class-incremental learning that uses virtual outliers to tighten decision boundaries, reducing confusion between classes from different tasks. Experimental results demonstrate the role of representation learning, which focuses on reducing overlapping between class representations. 

(4) \textbf{S-Prompt++} \citep{wang2022sprompts}: S-Prompt was originally proposed for domain-incremental learning, training a separate prompt and classifier head for each task. During evaluation, it infers the domain ID using the nearest centroid from K-Means applied to the training data. To adapt S-Prompt to class-incremental learning (CIL), S-Prompt++ uses a common classifier head for all tasks. However, it shares limitations with DualP, such as efficient learning of the classification head and predicting appropriate prompts during testing.

(5) \textbf{CODA-Prompt} \citep{smith2022coda}: This prompt-based approach uses task-specific learnable prompts for each task. Similar to L2P, CODA employs a pool of prompts and keys, computing a weighted sum from these prompts to generate the real prompt. The weights are based on the cosine similarity between queries and keys. To avoid task prediction at the end of the task sequence, the weighted sum always considers all prompts. CODA improves over DualP and L2P by optimizing keys and prompts simultaneously, but it still hasn't addressed the drawbacks mentioned for DualP.

(6) \textbf{HiDe-Prompt} \citep{wang2023hierarchical}: a recent SOTA prompt-based method that decomposes learning CIL into 3 modules: a task inference, a within-task predictor and a task-adaptive predictor. The second module trains prompts for each task with a contrastive regularization that tries to push features of new tasks away from prototypes of old ones. To predict task identity, it trains a classification head on top of the pretrained ViT. TAP is similar to a fine-tuning step that aims to alleviate classifier bias using the Gaussian distribution of all classes seen so far. However, this method does not declare the relationship between data during training, thereby missing the opportunity to improve model performance.
 
(7) \textbf{CPP} \citep{li2024steering}: This recent SOTA also uses a contrastive constraint to control features of all tasks so far during representation learning and achieves roughly equivalent performance to HiDE on the same settings. Nevertheless, this method still has the advantages that we pointed out in HiDE, which we propose to address in our work.
% CPP cho m nhe. T quen bai nay roii:v

\subsection{Metrics}
\label{metrics}
% We use the final average accuracy (FAA), and final forgetting measure (FFM).
% Specifically, we define the accuracy on the $i$-th task after learning the $t$-th task as $A_{i,t}$, and define the average accuracy of all seen tasks as $AA_{t} = \frac{1}{t}\Sigma_{i=1}^{t}A_{i,t}$. After learning all $T$ tasks, we report $\text{FAA} = AA_{T}$, and $\text{FFM} = \frac{1}{T - 1}\Sigma_{i=1}^{T - 1}\max_{t \in \{1, \dots, T - 1\}}{(A_{i,t} - A_{i,T})}$. FAA is the primary metric to evaluate the final performance of continual learning, and FFM serves as a measure of catastrophic forgetting.

In our study, we employed two key metrics: the Final Average Accuracy (FAA) and the Final Forgetting Measure (FFM). To define these, we first consider the accuracy on the \(i\)-th task after the model has been trained up to the \(t\)-th task, denoted as \(A_{i,t}\). The average accuracy of all tasks observed up to the \(t\)-th task is calculated as \(AA_{t} = \frac{1}{t}\sum_{i=1}^{t}A_{i,t}\). Upon the completion of all \(T\) tasks, we report the Final Average Accuracy as \(\text{FAA} = AA_{T}\). Additionally, we calculate the Final Forgetting Measure, defined as \(\text{FFM} = \frac{1}{T - 1}\sum_{i=1}^{T - 1}\max_{t \in \{1, \ldots, T - 1\}}{(A_{i,t} - A_{i,T})}\). The FAA serves as the principal indicator for assessing the ultimate performance in continual learning models, while the FFM evaluates the extent of catastrophic forgetting experienced by the model.

\subsection{Implementation Details}
\label{Implementation}

 % In our method, we set the contrastive regularization weight $\beta=0.1$, the parameter for prompt construction $\eta=0.99$
Our implementation basically aligns with the methodologies employed in prior research \cite{wang2022learning_l2p, wang2022dualprompt, smith2022coda}. Specifically, our framework incorporates the use of a pretrained Vision Transformer (ViT-B/16) as the backbone architecture. For the optimization process, we utilized the Adam optimizer, configured with hyper-parameters $\beta_{1}$ set to $0.9$ and $\beta_{2}$ set to $0.999$. The training process was conducted using batches of $24$ samples, and a fixed learning rate of $0.03$ was applied across all models except for CODA-Prompt. For CODA-Prompt, we employed a cosine decaying learning rate strategy, starting at $0.001$. Additionally, a grid search technique was implemented to determine the most appropriate number of epochs for effective training. Regarding the pre-processing of input data, images were resized to a standard dimension of $224 \times 224$ pixels and normalized within a range of $[0, 1]$ to ensure consistency in input data format. The detailed values of the parameters can be found in our \href{https://anonymous.4open.science/r/HierC-089B/}{source code}.

In Table \ref{table:main_results} of the main paper, the results of L2P, DualPrompt, S-Prompt++, CODA-Prompt, and HiDe-Prompt on Split CIFAR-100 and Split ImageNet-R are taken from \citep{wang2023hierarchical}. Their results on the other two datasets are produced from the official code provided by the authors. For CPP and OVOR, the reported results are also reproduced from their official code. It's worth noting that the reported forgetting of HiDE is reproduced from their official code. 
% Besides, the results of ADaM and RanPAC on Split CIFAR-100, Split ImangeNet-R, and CUB are also taken directly from RanPAC paper \citep{mcdonnell2023ranpac}, while the remaining is produced by their official code.

\section{Using LLMs (i.e., ChatGPT, DeepSeek, etc.,) to build tree-like taxonomy during a sequence of tasks, incrementally}
\label{chatGPT}

We use the following prompt structure to generate the taxonomies:

\texttt{Given the label list: \textcolor{blue}{['$\cdots$']}, provide me the taxonomy from this list, based on their origin, type, and shape, so that the image encoders can recognize their images.}

% Labels list of CIFAR100: 

% \texttt{[
%     "apple",
%     "aquarium\_fish",
%     "baby",
%     "bear",
%     "beaver",
%     "bed",
%     "bee",
%     "beetle",
%     "bicycle",
%     "bottle",
%     "bowl",
%     "boy",
%     "bridge",
%     "bus",
%     "butterfly",
%     "camel",
%     "can",
%     "castle",
%     "caterpillar",
%     "cattle",
%     "chair",
%     "chimpanzee",
%     "clock",
%     "cloud",
%     "cockroach",
%     "couch",
%     "crab",
%     "crocodile",
%     "cup",
%     "dinosaur",
%     "dolphin",
%     "elephant",
%     "flatfish",
%     "forest",
%     "fox",
%     "girl",
%     "hamster",
%     "house",
%     "kangaroo",
%     "keyboard",
%     "lamp",
%     "lawn\_mower",
%     "leopard",
%     "lion",
%     "lizard",
%     "lobster",
%     "man",
%     "maple\_tree",
%     "motorcycle",
%     "mountain",
%     "mouse",
%     "mushroom",
%     "oak\_tree",
%     "orange",
%     "orchid",
%     "otter",
%     "palm\_tree",
%     "pear",
%     "pickup-truck",
%     "pine\_tree",
%     "plain",
%     "plate",
%     "poppy",
%     "porcupine",
%     "possum",
%     "rabbit",
%     "raccoon",
%     "ray",
%     "road",
%     "rocket",
%     "rose",
%     "sea",
%     "seal",
%     "shark",
%     "shrew",
%     "skunk",
%     "skyscraper",
%     "snail",
%     "snake",
%     "spider",
%     "squirrel",
%     "streetcar",
%     "sunflower",
%     "sweet\_pepper",
%     "table",
%     "tank",
%     "telephone",
%     "television",
%     "tiger",
%     "tractor",
%     "train",
%     "trout",
%     "tulip",
%     "turtle",
%     "wardrobe",
%     "whale",
%     "willow\_tree",
%     "wolf",
%     "woman",
%     "worm"
% ]}

\vspace{4mm}
Example output, when the list \textcolor{blue}{['$\cdots$']} is \texttt{["leopard", "rabbit", "mouse", "camel", "trout", "aquarium\_fish", "snake", "rose", "lawn\_mower", "bottle"]}:

\vspace{3mm}

\begin{lstlisting}
taxonomy = {
    "Natural": {
        "Animals": {
            "Mammals": {
                "Four-legged": ["leopard", 
                "rabbit", "mouse", "camel"]
            },
            "Aquatic": ["trout", 
            "aquarium_fish"],
            "Reptiles": ["snake"]
        },
        "Plants": {
            "Flowers": ["rose"]
        }
    },
    "Man-Made": {
        "Objects": {
            "Tools": ["lawn_mower"],
            "Containers": ["bottle"]
        }
    }
\end{lstlisting}

\begin{table*}[ht]
    % \vspace{-0.5cm}
    \centering
    
     % \vspace{-0.2cm}
	\smallskip
      \renewcommand\arraystretch{1.5}
    \small{
	\resizebox{0.8\textwidth}{!}{ 
        \centering
	\begin{tabular}{lcccccccccccc}
	 \hline
      \multirow{2}{*}{LLMs used to build taxonomies} & \multicolumn{2}{c}{Split CIFAR-100} & \multicolumn{2}{c}{Split Imagenet-R} &
      \multicolumn{2}{c}{Split CUB-200}  \\
      \cmidrule(lr){2-3} \cmidrule(lr){4-5} \cmidrule(lr){6-7}  
        & Sup-21K & iBOT-21K & Sup-21K & iBOT-21K & Sup-21K & iBOT-21K \\
        \hline
        \textit{Baseline (HiDE-Prompt)} & \textit{92.61} & \textit{93.02} & \textit{75.06} & \textit{70.83} & \textit{86.56} & \textit{78.23} \\ 
        \hline
        % \hdashline
        Llama-3-70b-Groq & 93.58 & \underline{93.82} & \textbf{77.01} & \textbf{73.12} & \textbf{88.00} & \textbf{79.05}\\
        GPT-4o-Mini & \textbf{93.94} & \textbf{94.01} & \underline{76.41} & \underline{72.12} & \underline{87.93} & \underline{79.02}\\
        Gemini-1.5-Pro & 93.35 & 93.43 & 75.76 & 71.42 & 87.02 & 78.43\\
        DeepSeek-R1 & 93.56 & 93.62 & 76.24 & 71.93 & 87.65 & 78.72\\
        Mistral-Medium & 93.12 & 93.35 & 75.54 & 71.12 & 87.00 & 78.22\\
        Claude-3.5-Haiku & \underline{93.65} & 93.62 & 76.12 & 71.98 & 87.83 & 78.75\\
       \hline
       Average results of LLMs & $93.53_{\pm 0.28}$	& $93.64_{\pm 0.24}$ &	$76.18_{\pm 0.52}$ & $71.95_{\pm 0.69}$ &	$87.57_{\pm 0.45}$	& $78.8_{\pm 0.33}$\\
       \hline
	\end{tabular}}}
        \caption{Performance comparison when using different LLMs to generate the corresponding taxonomy} 
	\label{table:diff_llms}
	\vspace{-0.2cm}
\end{table*}

Below is an example of generated taxonomies for each task of Split-CIFAR100:

\begin{lstlisting}
T1 = {
    "Natural": {
        "Animals": {
            "Mammals": {
                "Four-legged": ["leopard", 
                "rabbit", "mouse", "camel"]
            },
            "Aquatic": ["trout", 
            "aquarium_fish"],
            "Reptiles": ["snake"]
        },
        "Plants": {
            "Flowers": ["rose"]
        }
    },
    "Man-Made": {
        "Objects": {
            "Tools": ["lawn_mower"],
            "Containers": ["bottle"]
        }
    }
}

T2 = {
    "Natural": {
        "Animals": {
            "Mammals": {
                "Four-legged": ["leopard", 
                "rabbit", "mouse", "camel", 
                "otter"]
            },
            "Aquatic": ["trout", 
            "aquarium_fish", 
                        "shark", "seal", 
                        "lobster"],
            "Reptiles": ["snake"]
        },
        "Plants": {
            "Flowers": ["rose", "tulip"],
            "Trees": ["palm_tree"]
        }
    },
    "Man-Made": {
        "Objects": {
            "Tools": ["lawn_mower"],
            "Containers": ["bottle", "bowl"]
        },
        "Vehicles": {
            "Wheeled": ["motorcycle"]
        },
        "Structures": {
            "Buildings": ["skyscraper", 
                             "house"]
        }
    }
}

T3 = ...
}
\end{lstlisting}

The taxonomies for other datasets are available in our \href{https://anonymous.4open.science/r/HierC-089B/}{source code}.

\section{Additional experiments} \label{additional_exp}

\begin{figure*}
    \centering
    % Row 1
    \begin{subfigure}[t]{0.7\linewidth}
        \centering
        \includegraphics[width=\linewidth]{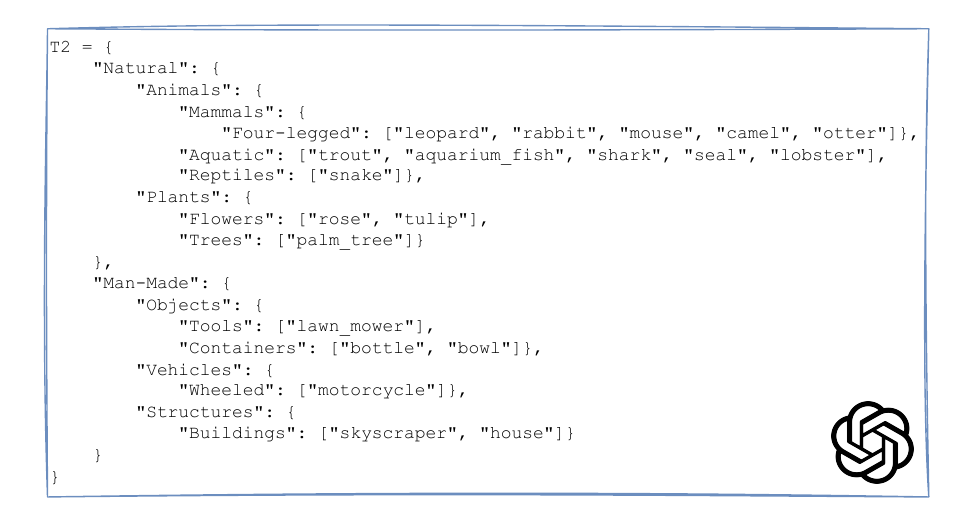}
        \vspace{-8mm}
        \caption{GPT-4o-Mini}
        \label{fig:subfig1}
    \end{subfigure}
    % \vspace{1em}
    \begin{subfigure}[t]{0.7\linewidth}
        \centering
        \includegraphics[width=\linewidth]{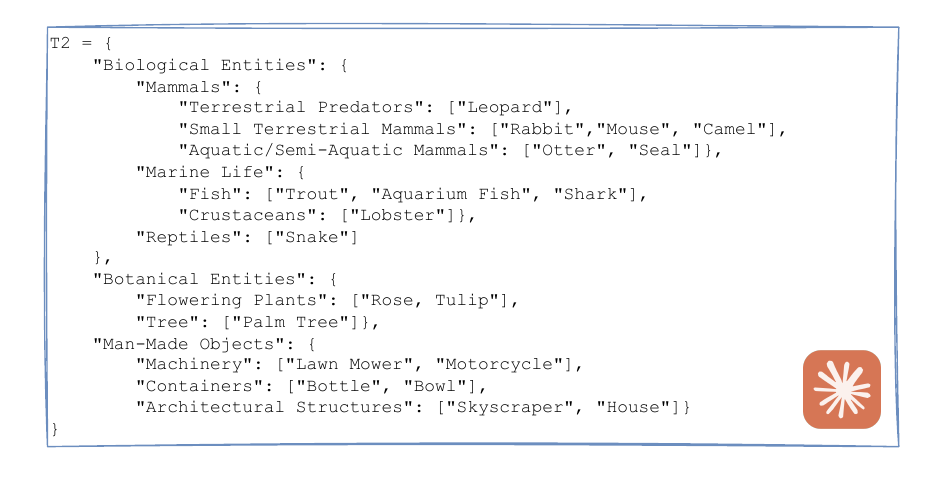}
        \vspace{-8mm}
        \caption{Claude-3.5-Haiku}
        \label{fig:subfig2}
    \end{subfigure}

    % \vspace{1em} % Space between rows

    % Row 2
    \begin{subfigure}[t]{0.7\linewidth}
        \centering
        \includegraphics[width=\linewidth]{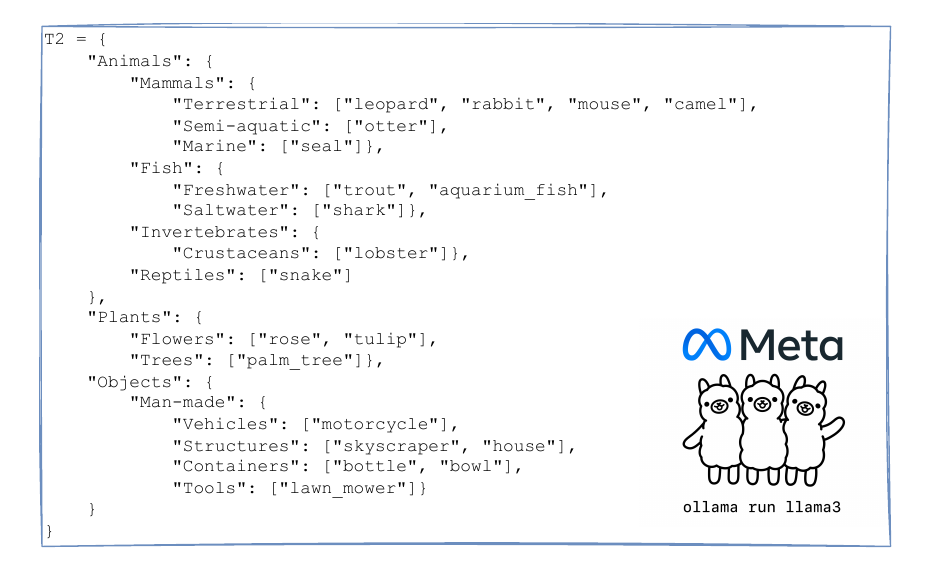}
        \vspace{-8mm}
        \caption{Llama-3-70b-Groq}
        \label{fig:subfig3}
    \end{subfigure}

    \caption{Taxonomy samples when using different LLMs (Llama, GPT, Claude), in \textit{Task 2}, with the label list \texttt{["leopard", "rabbit", "mouse", "camel", "trout", "aquarium\_fish", "snake", "rose", "lawn\_mower", "bottle"]}}
    \label{fig:llm1}
\end{figure*}

\begin{figure*}
    \centering
    % Row 1
    \begin{subfigure}[t]{0.7\linewidth}
        \centering
        \includegraphics[width=\linewidth]{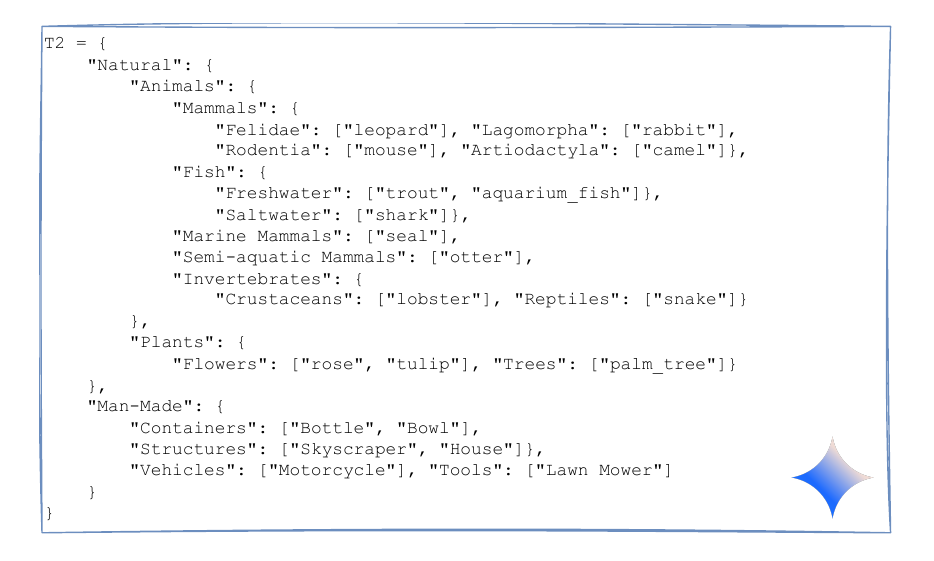}
        \vspace{-8mm}
        \caption{Gemini-1.5-Pro}
        \label{fig:subfig4}
    \end{subfigure}
    % \vspace{1em}
    \begin{subfigure}[t]{0.7\linewidth}
        \centering
        \includegraphics[width=\linewidth]{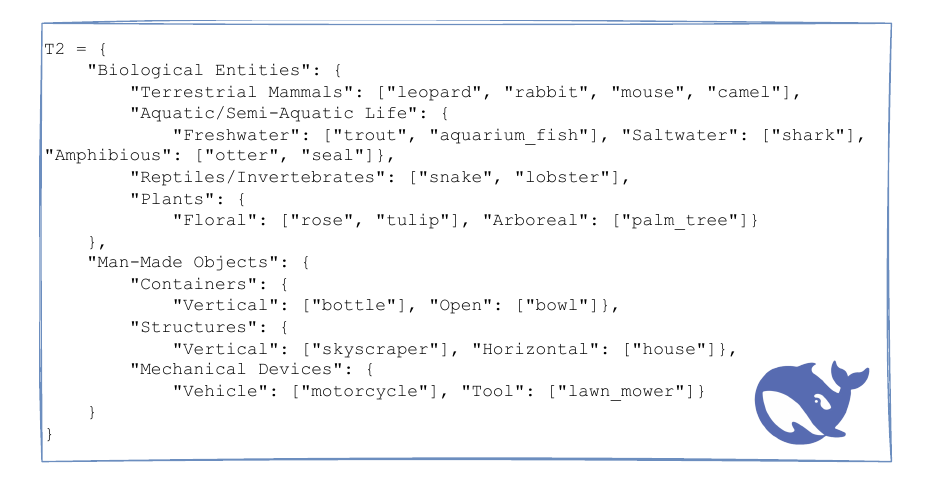}
        \vspace{-8mm}
        \caption{DeepSeek-R1}
        \label{fig:subfig6}
    \end{subfigure}

    % \vspace{1em} % Space between rows

    % Row 2
    \begin{subfigure}[t]{0.7\linewidth}
        \centering
        \includegraphics[width=\linewidth]{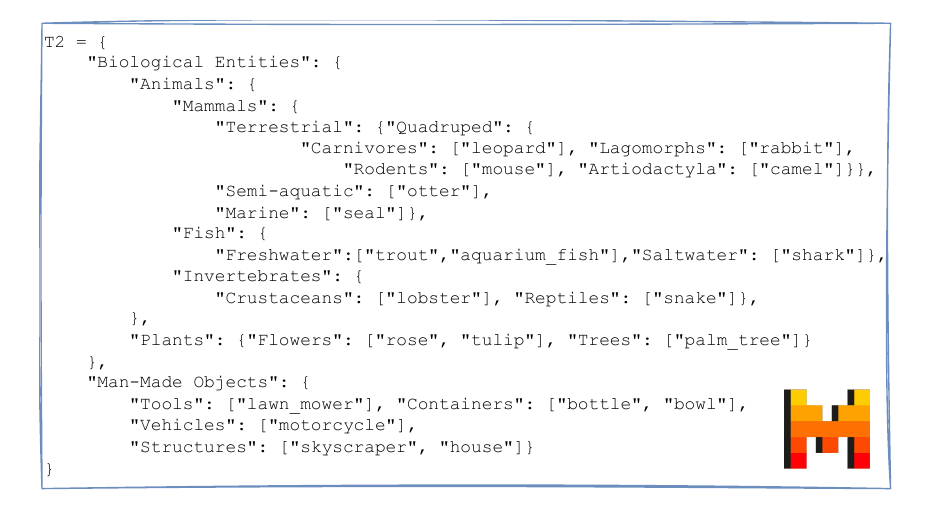}
        \vspace{-8mm}
        \caption{Mistral-Medium}
        \label{fig:subfig5}
    \end{subfigure}

    \caption{Taxonomy samples when using different LLMs (Gemini, DeepSeek, Mistral), in \textit{Task 2}, with the label list \texttt{["leopard", "rabbit", "mouse", "camel", "trout", "aquarium\_fish", "snake", "rose", "lawn\_mower", "bottle"]}}
    \label{fig:llm2}
\end{figure*}

\subsection{How do different hierarchical taxonomy structures affect model performance?}
\label{many_llms}
% Table . Figure \ref{fig:llm1}, \ref{fig:llm2}
Table \ref{table:diff_llms} shows the performance of models when using the most advanced LLMs to support the process of building label-based hierarchical taxonomies. Although the output samples presented in Figures \ref{fig:llm1} and \ref{fig:llm2} indicate some notable differences in the approaches as well as the final results of the corresponding LLMs' trees constructed, the numerical results show that there is not much difference when using these different LLMs. 
This may stem from the sufficiently good knowledge and semantic capabilities of these powerful models. That is, despite certain differences, most class labels are appropriately organized into their corresponding leaf groups. Furthermore, the involvement of our  OT-based technique once again helps to modify the constraint level of the taxonomy-based strategy, resulting in not much difference when using these LLMs.
Overall, Llama-3-70B-Groq and GPT-4o-Mini are the two models that yield the best results.

\subsection{Effect of the temperature $\tau$ in the final objective function}

In each of our experiments, we used a common temperature value $\tau$ for the two supervised contrastive loss functions (i.e., $\mathcal{L}_g$ and $\mathcal{L}_{all}$). Table \ref{table:tau} below shows the model performance when this value varies on Split-CIFAR-100 and Split-Imagenet-R. The results indicate that, in general, $\tau=0.1$ is the optimal value achieved.

% \begin{table}[ht]
%     % \vspace{-0.5cm}
%     \centering
     
%      \vspace{-0.2cm}
% 	\smallskip
%       \renewcommand\arraystretch{1.4}
%       % \setlength\tabcolsep{2.5pt}
%     \small{
% 	\resizebox{0.5\textwidth}{!}{ 
%         \centering
% 	\begin{tabular}{lccccc}
% 	 \hline
%       % \multirow{2}{*}{\,\,\,\,\,\,\,\,\,\,\,\,\,\,\,\, Dataset} & \multicolumn{5}{c}{Number of steps} \\
%       Dataset 
%         & $100$ & $200$ & $500$ & $1000$ & $5000$ \\
%         \hline
%        Split CIFAR-100 & 93.29 & 93.82 & \textbf{93.94} & 93.55 & 93.53\\ 
%        Split ImageNet-R & 75.66 & 76.21 & 76.62 & \textbf{77.01} & 76.84 \\ 
%        \hline
% 	\end{tabular}
% 	} 
% }
	
% 	% \vspace{-0.56cm}
%     \caption{Performance when $\delta$ (used to compute \textit{"the weight matrix"} in Eq.\ref{deltaaa}) varies}
%     \label{table:delta}
% \end{table}

\begin{table}[ht]
    % \vspace{-0.5cm}
    \centering
     
     \vspace{-0.2cm}
	\smallskip
      \renewcommand\arraystretch{1.4}
    \small{
	\resizebox{\linewidth}{!}{ 
        \centering
	\begin{tabular}{lccccc}
	 \hline
      % \multirow{2}{*}{\,\,\,\,\,\,\,\,\,\,\,\,\,\,\,\, Dataset} & \multicolumn{5}{c}{Number of steps} \\
      Dataset 
        & $0.05$ & $0.1$ & $0.3$ & $0.5$ & $0.7$  \\
        \hline
       Split CIFAR-100 & 90.04 & \textbf{93.94} & 91.42 & 90.15 & 88.81  \\ 
       Split ImageNet-R & 74.98 & \textbf{77.01} & 75.22 & 73.01 & 70.96 \\ 
       \hline
	\end{tabular}
	} 
}
	
	% \vspace{-0.56cm}
    \caption{Performance when the temperature factor $\tau$ (used in Eq.\ref{uuuu}) varies}
    \label{table:tau}
\end{table}

\subsection{The effectiveness of our OT-based technique}
\label{ot_onlyy}

\begin{table*}[ht]
    % \vspace{-0.5cm}
    \centering
     
     \vspace{-0.2cm}
	\smallskip
      \renewcommand\arraystretch{1.4}
    \small{
	\resizebox{.95\textwidth}{!}{ 
        \centering
	\begin{tabular}{lccccc}
	 \hline
      % \multirow{2}{*}{\,\,\,\,\,\,\,\,\,\,\,\,\,\,\,\, Dataset} & \multicolumn{5}{c}{Number of steps} \\
      Dataset & Split CIFAR-100 & Split ImageNet-R & 5-Dataset & Split CUB-200 \\

        % & $100$ & $200$ & $500$ & $1000$ & $5000$ \\
        \hline
       % Split CIFAR-100 & 93.29 & 93.82 & \textbf{93.94} & 93.55 & 93.53\\ 
       % Split ImageNet-R & 75.66 & 76.21 & 76.62 & \textbf{77.01} & 76.84 \\ 
       (A) Baseline (training tasks independently, $\mathcal{L}_{CE}$ only) & 87.52 & 70.55 & 91.48 & 81.02\\
       \textit{(B) Using OT-trick w/o taxonomies} & \textit{93.04} & \textit{75.69} & \textit{94.12} & \textit{86.85} \\
       (C) Using taxonomy w/o OT-based strategy & \underline{93.31} & \underline{75.72} & \underline{94.42} & \underline{87.18} \\
       (D) Ours (both taxonomy and OT-based strategy) & \textbf{93.94} & \textbf{76.41} & \textbf{95.02} & \textbf{87.93}\\
       \hline
	\end{tabular}
	} 
}
	
	% \vspace{-0.56cm}
    \caption{The efficiency of our OT-based technique}
    \label{table:ot_effect}
\end{table*}

This section provides an additional perspective on the effectiveness of our OT-based technique (Section \ref{ot_tricky}). Instead of demonstrating the effectiveness of this technique in reinforcing the strength of the taxonomy strategy as mentioned in the main paper, we conduct experiments to show its effectiveness when applied to conventional supervised contrastive loss. That is, at this point, the model's objective function is as follows:
\begin{align}
    \mathcal{L} = \mathcal{L}_{CE} + \mathcal{L_G}
    =  \mathcal{L}_{CE} + \mathcal{L}^{'}_{all}
    \label{loss_new}
\end{align}
where $\mathcal{L}^{'}_{all}$ is the new version of $\mathcal{L}_{all}$ after applying our OT-based strategy:
\begin{align}
    \displaystyle \mathcal{L}^{'}_{all}(\cdot) &= -\log \sum_{\vx'\in X_{1...t}|y_{\vx'}=y_\vx} \frac{u(\vz_x \cdot  \vz_{x'})}{\sum_{\bar{x} \in X_{1...t}} \textcolor{blue}{\bm{\gamma_{y_xy_{\Tilde{x}}}}}u(\vz_x \cdot  \vz_{\bar{x}})
    }
\end{align}

Based on this, we conducted the related experiments and presented the results in Table \ref{table:ot_effect}. The data show that the performance of the model when using the objective function in Eq.\ref{loss_new} \textit{(B)} is nearly equivalent to that of using our taxonomy strategy w / without the OT technique \textit{ (C)}, helping to improve performance by almost 6\% compared to training tasks independently \textit{ (A)}. The reason that \textit{(B)} has not surpassed \textit{(C)} may be because \textit{(B)} is only leveraging prior information from vision pretrained models, while \textit{(C)} also incorporates insights from LLMs regarding the relationships between data classes. The effectiveness of this collaboration is demonstrated when our fully proposed method \textit{(D)} delivers superior results.

\subsection{Model performance when $\alpha$, $\beta$ vary}

\begin{figure}[ht]
    \centering
    \vspace{-3mm}
    \includegraphics[width=0.8\linewidth]{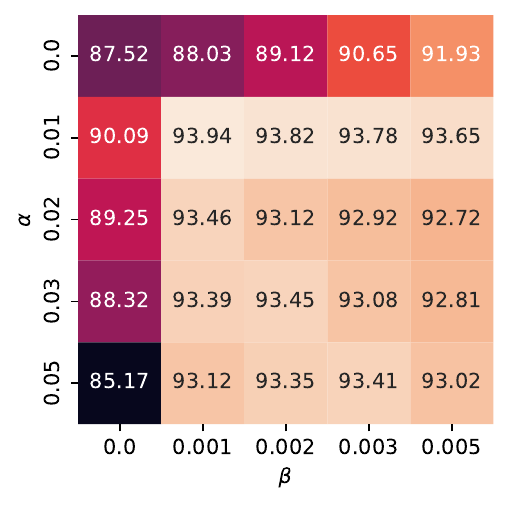}
    \vspace{-5mm}
    \caption{Varying the value of $\alpha, \beta$ (Split-CIFAR-100)}
    \label{fig:alpha_beta_cifar}
\end{figure}

\begin{figure}[ht]
    \centering
    \includegraphics[width=0.8\linewidth]{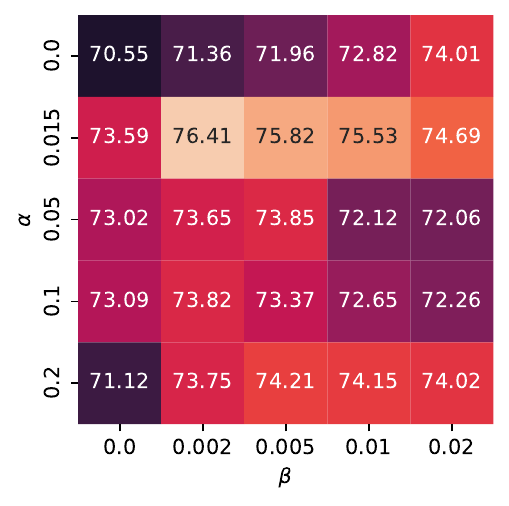}
    \vspace{-5mm}
    \caption{Varying the value of $\alpha, \beta$ (Split-Imagenet-R)}
    \label{fig:alpha_beta_imr}
    \vspace{-3mm}
\end{figure}

Figures \ref{fig:alpha_beta_cifar} and \ref{fig:alpha_beta_imr} show the changes in model performance when varying the values of $\alpha$ and $\beta$ in the objective function of the proposed method (Equation \ref{final_loss}). Accordingly, the roles of $\mathcal{L}^{'}_{g}$ and $\mathcal{L}_{all}$
  are clearly demonstrated and consistent with the insights obtained from the Split-CUB-200 dataset mentioned earlier in the main paper. When either of these quantities is omitted, model performance decreases significantly

\subsection{More verification about the alignment between taxonomy and visual space}

\begin{figure*}[!ht]
    \centering
    \begin{minipage}{0.55\textwidth}
        \centering
        \subcaptionbox{Label-based taxonomy\label{fig:sub1}}{\includegraphics[width=\linewidth]{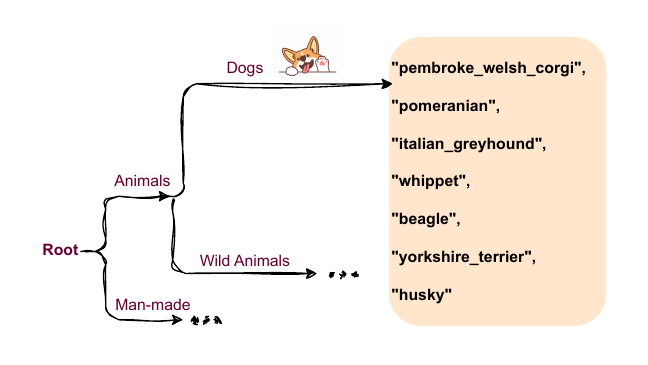}} % Replace with your image path
    \end{minipage}
    \begin{minipage}{0.4\textwidth}
        \centering
        \subcaptionbox{Visual Features. The numbers indicate the mean positions of corresponding classes. \label{fig:sub2}}{\includegraphics[width=\linewidth]{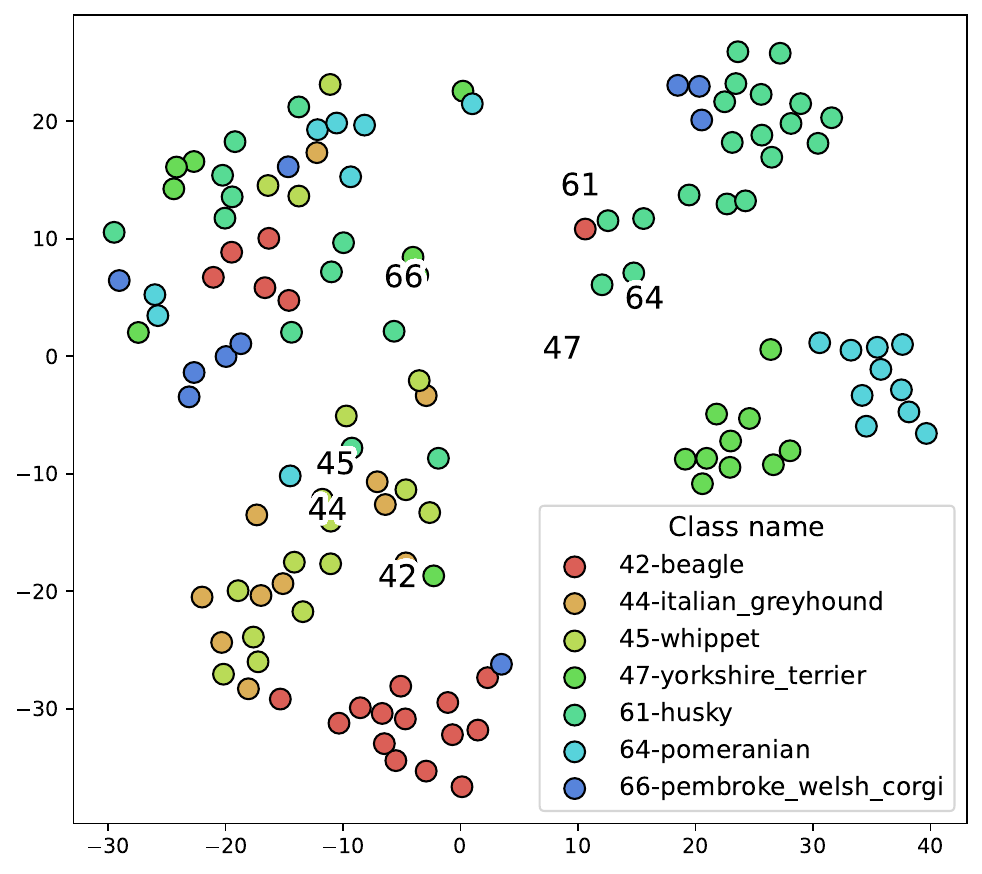}} % Replace with your image path
    \end{minipage}

    \begin{minipage}{0.8\textwidth}
        \centering
        \subcaptionbox{Corresponding real images\label{fig:sub3}}{\includegraphics[width=\linewidth]{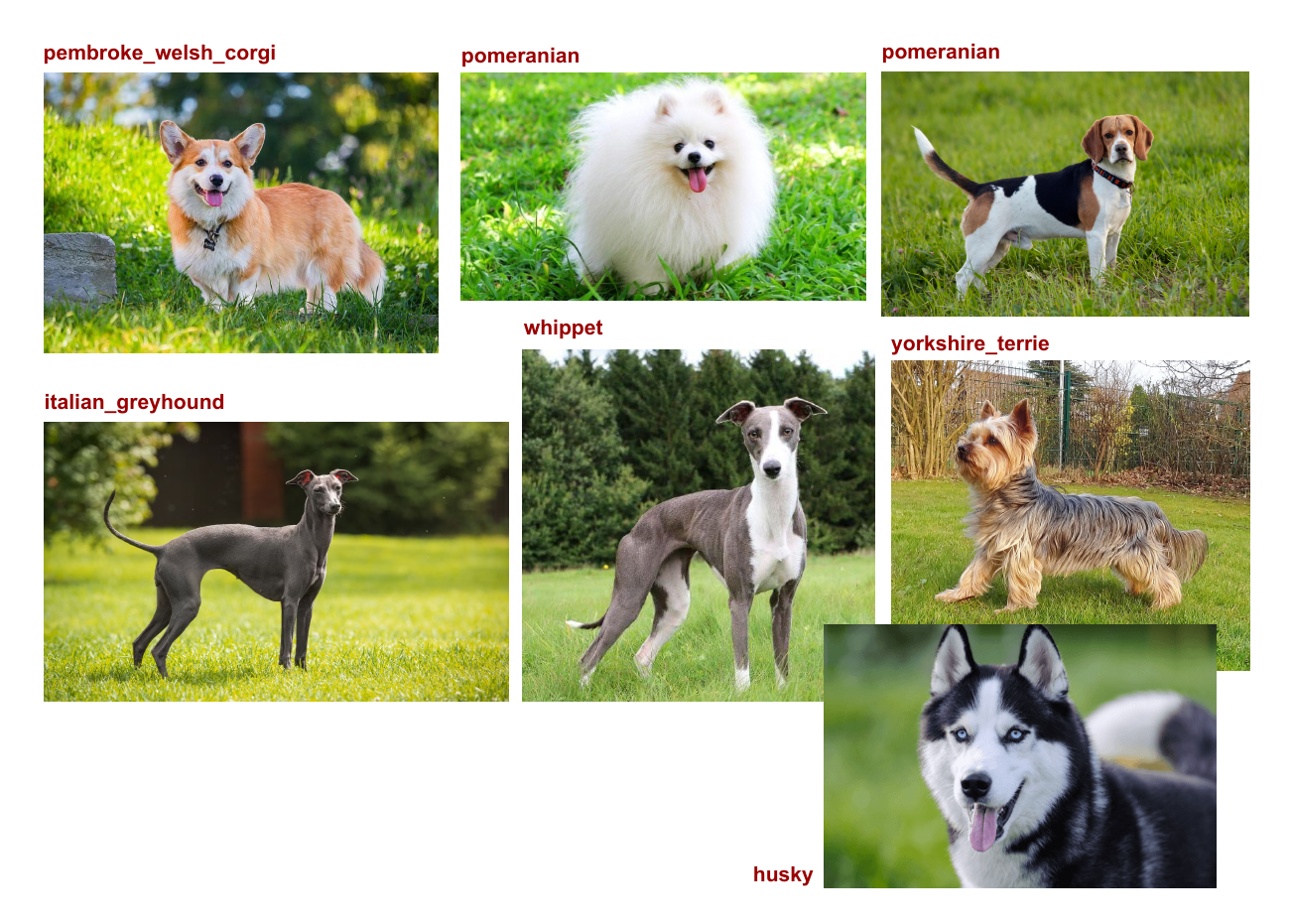}} % Replace with your image path
    \end{minipage}
    % \begin{minipage}{0.4\textwidth}
    %     \centering
    %     \subcaptionbox{Caption 4\label{fig:sub4}}{\includegraphics[width=\linewidth]{image4.jpg}} % Replace with your image path
    % \end{minipage}

    % \begin{minipage}{0.4\textwidth}
    %     \centering
    %     \subcaptionbox{Caption 5\label{fig:sub5}}{\includegraphics[width=\linewidth]{image5.jpg}} % Replace with your image path
    % \end{minipage}
    % \begin{minipage}{0.4\textwidth}
    %     \centering
    %     \subcaptionbox{Caption 6\label{fig:sub6}}{\includegraphics[width=\linewidth]{image6.jpg}} % Replace with your image path
    % \end{minipage}

    \caption{Relationship between visual features and taxonomies (Split-Imagenet-R)}
    \label{fig:relation_taxo_img}
    \vspace{-3mm}
\end{figure*}

To better demonstrate the intuition that there is a correlation between the label-based tree-like taxonomy and visual space, we provide visualization results corresponding to different datasets in Figure \ref{fig:relation_taxo_img}. The results show that classes within the same \textit{leaf group} often share many common visual characteristics and are likely to overlap in the latent space. Therefore, identifying these such groups on the taxonomies will provide us with a reference channel to determine easily confusable classes, thereby enhancing the models' representation learning.

\subsection{Discussion}
\paragraph{Limitation.} Despite our novel approach, the quality of the hierarchical taxonomy can depend on the quality of expert knowledge, thus affecting the model performance. For example, if similar image classes are not assigned to the same leaf group in this label-based taxonomy, the constraint we put on each such group may not perform as expected. Therefore, in our experiment, we utilize the latest power full LLMs to support this strategy, thereby mitigating the possibility of inappropriate organization which can harm performance. See Appendix \ref{many_llms}.

\paragraph{Potential impacts.} In this paper, we aim for Continual Learning for Image Classification tasks. However, this method may promise to effectively support more complex tasks in Computer Vision, such as Object Detection, Action Recognition, etc., in general. Furthermore, organizing concepts into groups may also hold promise for knowledge transfer strategies for CL settings, helping the models flexibly reinforce relevant knowledge according to each specific domain, which we will reserve for future work.